\documentclass[11pt]{article}
\usepackage{fullpage}

\usepackage[utf8]{inputenc} 
\usepackage[T1]{fontenc}    
\usepackage{hyperref}       
\usepackage{url}            
\usepackage{booktabs}       
\usepackage{amsfonts}       
\usepackage{nicefrac}       
\usepackage{microtype}      
\usepackage{tcolorbox}
\usepackage{diagbox}

\usepackage{enumerate}
\usepackage[shortlabels]{enumitem}
\usepackage{algorithmic}
\usepackage{algorithm}
\usepackage{graphicx} 
\usepackage{subcaption}
\usepackage{amsmath}
\usepackage{amsthm}
\usepackage{amssymb}
\usepackage{tikz}
\usepackage{tablefootnote}
\usepackage{multirow}
\usepackage{enumerate}
\usepackage{color}
\usepackage{xcolor}
\usepackage{natbib}

\usetikzlibrary{arrows}

\allowdisplaybreaks[4]

\usepackage{mathrsfs}

\usepackage{hyperref}
\usepackage{bm,todonotes}

\allowdisplaybreaks

\newtheorem{theorem}{Theorem}[section]
\newtheorem{lemma}{Lemma}[section]

\newtheorem{definition}{Definition}[section]

\newtheorem{remark}{Remark}[section]

\newtheorem{axiom}{Axiom}

\hypersetup{
	colorlinks=true,
	filecolor=blue,
	citecolor = blue,
	urlcolor=cyan,
}

\usepackage{caption, subcaption}
 
\usepackage{amsmath,amsfonts,bm}

\def\1{\bm{1}}

\DeclareMathAlphabet{\mathsfit}{\encodingdefault}{\sfdefault}{m}{sl}
\SetMathAlphabet{\mathsfit}{bold}{\encodingdefault}{\sfdefault}{bx}{n}

\def\c{{\bf c}}

\def\v{{\bf v}}

\def\Xi{\boldsymbol{\xi}}

\def\Phi{\boldsymbol{\phi}}

\def\0{{\bf 0}}
\def\1{{\bf 1}}

\title{
Variance reduced Shapley value estimation for trustworthy data valuation
}

\author{
Mengmeng Wu\thanks{School of Management, Xi'an Jiaotong University; email: \texttt{mengmengwu@stu.xjtu.edu.cn}. } \\
	\and
Ruoxi Jia\thanks{Virginia Tech, Blacksburg; email: \texttt{ruoxijia@vt.edu}. } \\
	\and 
Changle Lin\thanks{China Aerospace Academy of Systems Science and Engineering, Institute for Interdisciplinary Information Core Technology; email: \texttt{changlelin@tsinghua.edu.cn}. } \\
	\and
Wei Huang\thanks{College of Business, Southern University of Science and Technology; email: \texttt{huangw7@sustech.edu.cn}. } \\
	\and
	Xiangyu Chang \thanks{Corresponding author: Center for Intelligent Decision-Making and Machine Learning, School of Management, Xi’an Jiaotong University; email: \texttt{xiangyuchang@xjtu.edu.cn}. } 
}
\begin{document}

\maketitle

\begin{abstract}%
Data valuation, especially quantifying data value in algorithmic prediction and decision-making, is a fundamental problem in data trading scenarios. 
The most widely used method is to define the data Shapley and approximate it by means of the permutation sampling algorithm. 
To make up for the large estimation variance of the permutation sampling that hinders the development of the data marketplace, we propose a more robust data valuation method using stratified sampling, named variance reduced data Shapley (VRDS for short). 
We theoretically show how to stratify, how many samples are taken at each stratum, and the sample complexity analysis of VRDS. 
Finally, the effectiveness of VRDS is illustrated in different types of datasets and data removal applications.
\end{abstract} 

\section{Introduction}~\label{sec:intro}
The emerging big data in all walks of life has become the driving force of technological and economic development~\citep{ghorbani2019data,huang2021toward}.
Various sectors such as finance and healthcare increasingly rely on individuals' data for predictions, decision-making, and generating business value, which promotes extensive data transactions~\citep{barua2012measuring}.
One of the most critical problems in data trading scenarios is data valuation.
We consider data trading scenarios in data markets based on machine learning models, such as DATABRIGHT~\citep{dao2018databright} and Sterling~\citep{2018sterling}.
The data value in this scenario is largely determined by its contribution to a specific machine learning model.
We focus on data valuation in supervised learning, which is one of the main pillars of machine learning. 
The core challenge is how to fairly evaluate the contribution of each data in the training set to the learning algorithm for a particular performance metric.

A natural way to handle the aforementioned issue is to treat each data as a player in a cooperative game. Then, the value of each player can be assessed through utility functions from a game-theoretic perspective~\citep{jia2019towards}.
The Shapley value~\citep{1953A}, a solution concept that egalitarian distributes both gains and costs to several participants in a coalition, has been generalized to evaluate the contribution of each datum in supervised learning~\citep{ghorbani2019data,jia2019efficient,jia2019towards,kwon2022beta,tang2021data}.
The Shapley value of a player is the average of its marginal contribution to all alliances that do not include itself. 
Since $2^{n-1}$ coalitions exclude it (assuming the number of players is $n$), the computational complexity of Shapley value is exponential.
Indeed, the exact computation of Shapley value is NP-hard in general~\citep{deng1994complexity}.
How to effectively approximate the Shapley value of each data in supervised learning is the key to its application.
The most widely used algorithm tackles this problem is the permutation sampling algorithm (also called Monte Carlo sampling)~\citep{ghorbani2019data,jia2019towards,tang2021data,vstrumbelj2014explaining,castro2009polynomial,strumbelj2010efficient,cohen2007feature}.
One first samples a random permutation of training data, then scans one by one from the first element to the last element in the permutation, calculates the marginal contribution of each element to the set of elements in front of it, and finally repeats the same procedure over multiple permutations and takes the average of all their marginal contributions as the approximation of Shapley values. 
We refer to Shapley value for data valuation as data Shapley. 

The permutation sampling gives an unbiased estimate of the data Shapley. 
However, it does not consider the impact of the cardinality of the training set on the model performance in machine learning.
When we sample different permutations, the cardinality of the training set used to calculate the marginal contribution of each data may be different, resulting in relatively significant variances (see Figure~\ref{FashionMNIST} for detailed discussions).
However, the reproducibility of data valuation is the key to building trust in data transactions, and minimizing the variance of estimation results is critical to data valuation.

Our research aims to figure out the deficiency mentioned above of permutation sampling to estimate data Shapley and provide a more robust data valuation approach that can reduce the variance of estimating data Shapley. Therefore, we propose the following research questions (RQs): \textit{What methods can be adopted to reduce the variance of evaluating data Shapley? What are their theoretical advantages and disadvantages? How to efficiently implement the proposed method in practice?} 
To answer these RQs, we are inspired by the so-called \textit{stratified sampling} to propose a novel data Shapley estimation method called \textit{Variance Reduced Data Shapley} (VRDS). 
Stratified sampling commonly divides the target population into several types or layers according to its attribute characteristics and then randomly selects samples from the layers~\citep{cochran1977sampling,lohr2021sampling}.
Stratified sampling increases the commonality of elements in each layer through classification and stratification, and it is easy to extract representative survey samples and further reduce the estimate variance. 
Moreover, we establish the total variance of stratified sampling for estimating data Shapley. 
To minimize the total variance, the optimal proportion of the number of sampled permutations per layer is manifested. 
The result is that the number of samples in each layer is directly proportional to the standard deviation of the marginal contribution in the corresponding layer. 
Since the actual standard deviation of each layer is unknown, how to effectively approximate the variance is a problem that must be considered.
We justify that the function of set cardinality can be used as an upper bound on the variance, and approximate the variance by it, and then obtain the number of samples of each layer.
Finally, the sample complexity analysis has been conducted for VRDS to demonstrate the theoretical property compared with the permutation sampling.

The rest of this paper is organized as follows. 
We discuss related works and contributions in Section \ref{sec:related_work}. Section \ref{sec:pre} introduces the data Shapley in supervised learning and presents the permutation sampling algorithm for estimating data Shapley. 
We propose the VRDS, including a stratified sampling algorithm and sample complexity analysis in Section \ref{sec:vrds}. 
The experimental study of VRDS is provided in Section \ref{sec:experiment}. Section \ref{sec:conclusion} summarizes the conclusions. 
All the technical proofs are presented in Appendix.

\section{Related Work and Contribution}\label{sec:related_work}

Data valuation or pricing has become a significant role in the digital economy. Thus, various data valuation schemes have been studied in the literature. 
As we all know, data product replication costs are meager, even close to zero in many scenarios, so the traditional pricing strategy can no longer be directly applied to data valuation~\citep{pei2020survey}.
In recent years, many articles have emerged to study different fields of data valuation, including arbitrage-free valuation~\citep{2014A,2019Towards}, revenue maximization valuation~\citep{2019Revenue}, fair and truthful valuation~\citep{ghorbani2019data,jia2019towards}, and privacy-preserving valuation~\citep{li2014pricing}.
This paper aims to study the fair and truthful valuation of data, that is, price each data fairly according to the impact of data on the model performance in supervised learning.

Shapley value was introduced in the classical game theory~\citep{1953A}. 
Recent studies have adopted the Shapley value to measure the importance of inputs to output in a specific system.
For example, suppose the specific system can be presented as a deterministic function. In that case, studies examine the impact of input variables on outputs by exploring their contribution to the variance of output, referred to as variance-based sensitivity analysis~\citep{owen2014sobol,song2016shapley,owen2017shapley,iooss2019shapley,broto2020variance,benard2022shaff,herin2022proportional,da2021basics}.
Several approximate methods have been proposed in this field, such as Monte Carlo algorithm~\citep{song2016shapley}, meta-model~\citep{iooss2019shapley}, double Monte Carlo~\citep{broto2020variance}, and Pick and Freeze estimators~\citep{broto2020variance}.
In contrast, there are significant differences between these papers and our work in data valuation. 
Firstly, they assess the impact of input variables or features on output, which differs from our consideration of the impact of samples on output. 
Secondly,  in variance-based sensitivity analysis, the functional relationship between input and output is assumed to be predetermined, but in the data valuation we study, the function is learned through training on samples.
Lastly, most of them aim to measure the importance of variables through variance decomposition, which is different from our goal of reducing the total variance of data value estimation.


Another research line assumes the system is the supervised learning model. 
This case studies the significance of features or data in supervised learning settings by analyzing their impact on the performance of the learning model
\citep{ghorbani2019data,jia2019towards,jia2019efficient,tang2021data,vstrumbelj2014explaining,strumbelj2010efficient,cohen2007feature}.
To mitigate the substantial computational expense of Shapley value, several approximate algorithms have been developed in these studies. 
These approximation approaches include the permutation sampling algorithm~\citep{ghorbani2019data, jia2019towards}, gradient Shapley algorithm~\citep{ghorbani2019data}, group testing-based algorithm~\citep{jia2019towards}. 
Although several approximation algorithms exist, the permutation sampling algorithm is the most frequently used and has been applied to data valuation~\citep{ghorbani2019data,jia2019towards,jia2019efficient,kwon2022beta,tang2021data}, feature attribution~\citep{strumbelj2010efficient,cohen2007feature}, cooperating game evaluation~\citep{vstrumbelj2014explaining,castro2009polynomial}.
Therefore, the permutation sampling algorithm has been confirmed as “the most useful baseline” for approximately computing Shapley value. 
Nonetheless, this algorithm suffers from significant variance resulting from random sampling, which limits its wide application (see Figure \ref{FashionMNIST} for discussions).
Our study concentrates on data valuation in supervised machine learning and seeks to address the limitations of the permutation sampling algorithm by utilizing a different method, namely stratified sampling.

The stratified sampling has been considered in the cooperative game theory.
\cite{Maleki2013bounding} use the stratified sampling to improve the bound of the approximation error. That is, they first scale the difference between Shapley and its estimated value to obtain the expression related to the sampling number of each layer. By minimizing this difference, they decide the sampling number of each layer and give the specific implementation method of stratified sampling.
\cite{2017improving} employ stratified sampling to estimate Shapley value, by reducing the variance of the estimations obtained by stratified sampling giving the formula of sampling number in every stratum. 
They consider that one of the sources of variance for each marginal contribution is the player and the order in which that player arrives, so they take two sources of variation when stratifying.  
\cite{2021Approximating} derive a concentration inequality that is tailored to stratified Shapley value estimation using sample variance information. 
Based on this error bound, they propose an online process of sequentially choosing  samples from the strata in order to minimize the estimate error.
Our research focuses more on how to apply the Shapley value to the data pricing market. 
In the data market scenario, the large variance brought by algorithms for estimating data Shapley is unacceptable. 
Therefore, we consider using the stratified sampling method, but unlike ~\cite{Maleki2013bounding}, we aim at the minimum variance of estimators from the algorithmic randomness.
The algorithm in \cite{2017improving} applies to situations where the variability of the marginal contributions depends greatly on each player’s arrival position and is also not suitable for our scenario. 
While the algorithm in \cite{2021Approximating} selects samples online according to the conclusion of the central inequality to reduce the estimation error, which increases the unnecessary calculation cost for the data markets and does not consider the variance of the estimated value.
Therefore, we propose a trustworthy data pricing approach specifically for data markets.

We summarize the contributions of this paper as follows:

\begin{itemize}
    \item Proposing a trustworthy data Shapley estimation method, VRDS, based on the stratified sampling. The optimal number of samples in each layer is determined by minimizing the variance of VRDS. We justify that the variance of VRDS is less than or equal to the variance estimated by the permutation sampling algorithm.

    \item Implementing a stratified sampling algorithm for estimating VRDS. We theoretically provide the sample complexity analysis of the algorithm. The theoretical result indicates that the sample number of proposed VRDS based on the stratified sampling has the same order with the permutation sampling up to a log factor for achieving an $(\epsilon,\delta)$-approximation.
    
    \item Designing expensive experimental studies on rich data sets to illustrate the effectiveness of VRDS. It is found that it can significantly reduce variance and effectively identify data quality.
\end{itemize}

\section{Preliminaries}\label{sec:pre}
In this section, we review Shapley value's concept and basic properties, based on which we set the framework for data valuation. 
We then introduce data Shapley for supervised learning and discuss the baseline permutation sampling algorithm to approximate the data Shapley.

\subsection{Shapley Value}
The Shapley value is a classic concept in cooperative game theory~\citep{1953A}.
A cooperative game is defined by a tuple $<N,U>$, where $N=\{1,2,\dots,n\}$ which denotes the set of all players, and $U:2^N \rightarrow \mathbb{R}$ is a map that assigns to each
coalition $S\subseteq N$ a real number $U(S)$ such that $U(\varnothing) = 0$. It represents the utility of the collaboration of the members in $S$.
The goal is to distribute the total income $U(N)$ among players according to each player's contribution to the cooperation.
Shapley value of player $i$ is defined as the average marginal contribution of $i$ to all coalition that excludes $i$
\begin{equation}\label{Shapley}
    \phi_i(U)=\frac{1}{n}\sum_{S\subseteq N\backslash\{i\},|S|=s}\frac{1}{\tbinom{n-1}{s}}[U(S\cup \{i\})-U(S)], \quad i=1,\dots,n,
\end{equation}
where $U(S\cup \{i\})-U(S)$ is the marginal contribution of player $i$ with respect to $S$. 

It has been justified that Shapley value is the unique solution for cooperative games satisfying the following axioms~\citep{1953A}, which prompts researchers to use it for data valuation~\citep{ghorbani2019data,jia2019towards}.

\begin{axiom}{(efficiency axiom)}
$\sum_{i\in N}\phi_i(U)=U(N)$.  
\end{axiom}

\begin{axiom}{(symmetry axiom)}
For players $i$ and $j$, if $U(S\cup \{i\})=U(S\cup \{j\})$ holds for all $S$, where $S\subseteq N$ and $i,j\notin S$, then $\phi_i(U)=\phi_j(U)$.  
\end{axiom}

\begin{axiom}{ (dummy axiom)}
 If $U(S\cup \{i\})=U(S)$ holds for all $S$, where $S\subset N$ and $i\notin S$, then $\phi_i(U)=0$.
\end{axiom}

\begin{axiom}{(additivity axiom)}
 For any pair of games $(N,U)$ and $(N,V)$, if $(U+V)(S)=U(S)+V(S)$ holds for all $S$, then $\phi(U+V)=\phi(U)+\phi(V)$. 
\end{axiom}

The efficiency axiom states that players would expect to distribute all utility of their coalition. 
The symmetry axiom means that if two players have the same marginal contribution to all coalitions excluding them, they should have the same value. 
The dummy axiom requires value assignment should be sensitive to the player's contribution to all coalitions. 
A null player's marginal contribution is zero to any coalition that does not include itself.
The additivity axiom can decompose a given utility function into an arbitrary sum of utility functions, and the value can be calculated respectively.
Shapley value satisfies the above four axioms, so it is considered as an egalitarian distribution of cooperative games. 
This prompts scholars to regard the process of data co-training the model as a game, regard each data as a player, measure each data's contribution to the model with Shapley value, and conduct data pricing based on it.

\subsection{Data Shapley for Supervised Learning}
Consider a dataset $D=\{(x_i,y_i)\}_{i=1}^n$ containing $n$ samples, where $x_i$ describes features of the $i$th instance and $y_i$ corresponds to its label.
We focus on the value of data in $D$.
Given a learning algorithm $\mathcal{A}$ and a performance metric $U$ to evaluate how well the model performs, we train it on a subset of $D$ denoted by $S$ and use a test set to compute the corresponding utility function $U(S)$.
If it is a regression model, the performance metrics can be MSE (Mean Squared Error), RMSE (Root Mean Squared Error), MAE (Mean Absolute Error), $R^2$ (R Squared), and so on. 
If it is a classification model, the performance indicators can include accuracy, F1 score, etc.
For example, if we consider a classification problem and select predictive accuracy as the utility function $U$, then $U(S)=\frac{1}{|T|}\sum_{j\in T}\mathrm{1}(y_j=\hat{g}_S(x_j))$, where $\hat{g}_S$ is the model trained on $S$ and $|T|$ is the sample size of $T$.
We aim to evaluate which data points significantly contribute to the corresponding evaluation metric $U$.
We can formulate the data valuation problem as a cooperative game by treating the data as players and the evaluation metric as a utility function. Then the Shapley value could be employed to evaluate the value of data.
The Shapley value of each data in supervised learning is referred to \textit{data Shapley}.

The main challenge in adopting data Shapley is its computational cost. 
The computational complexity of evaluating the exact data Shapley using Eq.(\ref{Shapley}) requires $\mathcal{O}(2^n)$, since it involves computing marginal contributions of all points to all sets.
In addition, for most algorithms, we need to train the machine learning model twice every time to calculate every marginal contribution, which is computationally expensive.

Many works of literature~\citep{ghorbani2019data,jia2019towards} have studied how to approximate data Shapley effectively.  
Up to now, the permutation sampling algorithm is a widely used baseline algorithm (see Algorithm \ref{RSA}). 
 Recall that Shapley value has an equivalent form (see \citet{ghorbani2019data}) as
\begin{equation}
\phi_i(U)=\mathbb{E}_{O\sim \Pi(N)}(U(P_i^O \cup \{i\})-U(P_i^O))=\frac{1}{n!}\sum_{O\in \Pi(N)}(U(P_i^O \cup \{i\})-U(P_i^O)),
\end{equation}
where $\Pi(N)$ is all permutations of $N$, and $P_i^O$ represents the set of predecessors of $i$ in the permutation $O$ . 
That is, if $O=\{i_1,i_2,\dots,i_k,i,i_{k+1},\dots,i_{n-1}\}$, $P_i^O=\{i_1,i_2,\dots,i_k\}$.
$P_i^O=\emptyset$ if $i$ is the first element.
According to the above formulation, we can consider that each permutation is uniformly sampled from $n!$ permutations with the probability $1/n!$. 
Intuitively, imagine all data are to be collected in a random order, and the marginal contribution of every data is to those already collected data.
If we average these marginal contributions over all possible data orders, we obtain $\phi_i$.
Thus, we can regard $\phi_i$ as the expectation of marginal contributions and estimate it by the sample mean.
The estimator $\hat{\phi}_i$ of $\phi_i$ is 
\begin{equation}
    \hat{\phi}_i(U)=\frac{1}{m}\sum_{O\in M}(U(P_i^O \cup \{i\})-U(P_i^O)),
\end{equation}
where $M$ represents permutation samples and $|M|=m$. 

\begin{algorithm}[!ht]
	\caption{Permutation Sampling Algorithm for Data Shapley}
	\label{RSA}
	\begin{algorithmic}[1]
	\REQUIRE
	Training data $D=\{(x_i,y_i)\}_{i=1}^n$, learning algorithm $\mathcal{A}$, performance score $U$, sampling number $m$.\\
    \ENSURE Data Shapley of training data: $\phi_1,\dots,\phi_n$.\\
    \STATE Initialize $\phi_i:=0$ for $i=1,\dots,n$;
	\FOR{$t=0$ to $m$}
	    \STATE $O^t$: random permutation of training data;\\
	    \STATE $u_0^t := U(\varnothing,\mathcal{A})$;\\
		\FOR{$j=1$ to $n$}
		\STATE $u_j^t := U(P_j^{O^t},\mathcal{A});$
		\STATE $\phi_{O^t[j]} := \frac{t-1}{t}\phi_{O^{t-1}[j]}+\frac{1}{t}(u_j^t-u_{j-1}^t);$
		\ENDFOR
	\ENDFOR
	\end{algorithmic}
\end{algorithm}

To evaluate the minimum sampling number to achieve a specific error level, we denote the following $(\epsilon,\delta)$-approximation of data Shapley.

\begin{definition}
$\hat{\phi}$ is an $(\epsilon,\delta)$-approximation to  $\phi$ if $\mathbb{P} [|\hat{\phi}- \phi|\leq \epsilon]\geq 1-\delta$. 
\end{definition}

Lemma \ref{permutation number} provides a lower bound on the sampling number $m$ in the permutation sampling algorithm to achieve an $(\epsilon,\delta)$-approximation.

\begin{lemma}
\label{permutation number}
Algorithm \ref{RSA} returns an $(\epsilon,\delta)$- approximation to data Shapley of one data if the sampling number of permutations $m$ satisfies $m \ge r^2\log (2/\delta)/2\epsilon^2$, where {$r$ is the range of the data's marginal contributions.}
In particular, if the utility function is the prediction accuracy, setting $m \ge \log (2/\delta)/2\epsilon^2$ is sufficient to achieve the $(\epsilon,\delta)$-approximation.
\end{lemma}

In machine learning, the range of utility functions is usually determined. For example, if we choose the utility function as the prediction accuracy, then $0\le U\le 1$. 
Therefore, the range of the marginal contributions of all data is $-1\le r\le 1$, which makes the result of Lemma \ref{permutation number} relatively simple and intuitive.
It is worth noting that this lemma has been proved by~\cite{Maleki2013bounding}. 
We refine the analysis for comparison purposes in subsequent sample complexity analyses of proposed VRDS.

\begin{figure}[htbp]
	\centering
	\begin{minipage}{0.49\linewidth}
		\centering
		\includegraphics[width=1\linewidth]{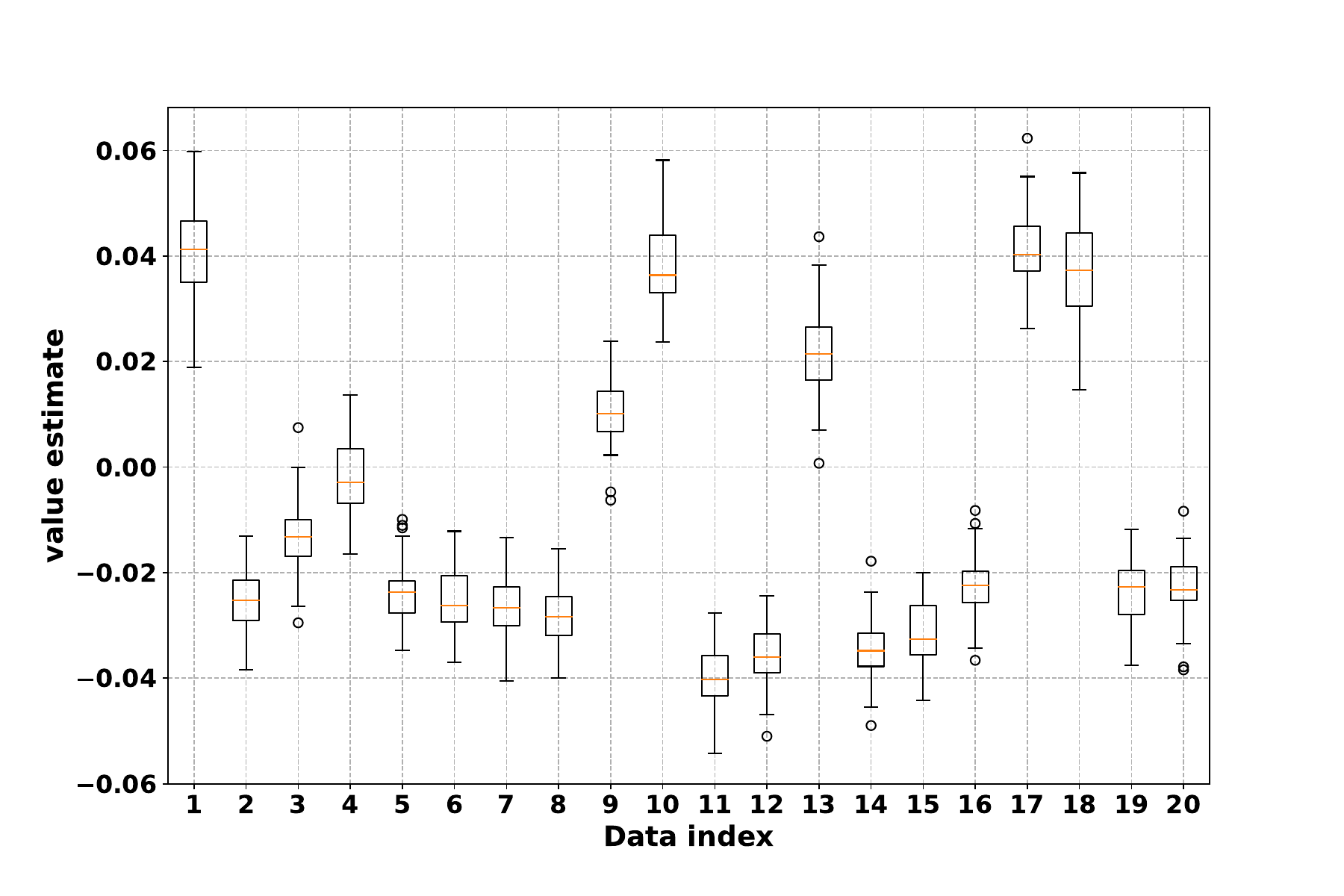}
            \subcaption{}
        \label{f11}
	\end{minipage}
	\begin{minipage}{0.49\linewidth}
		\centering		
  \includegraphics[width=1\linewidth]{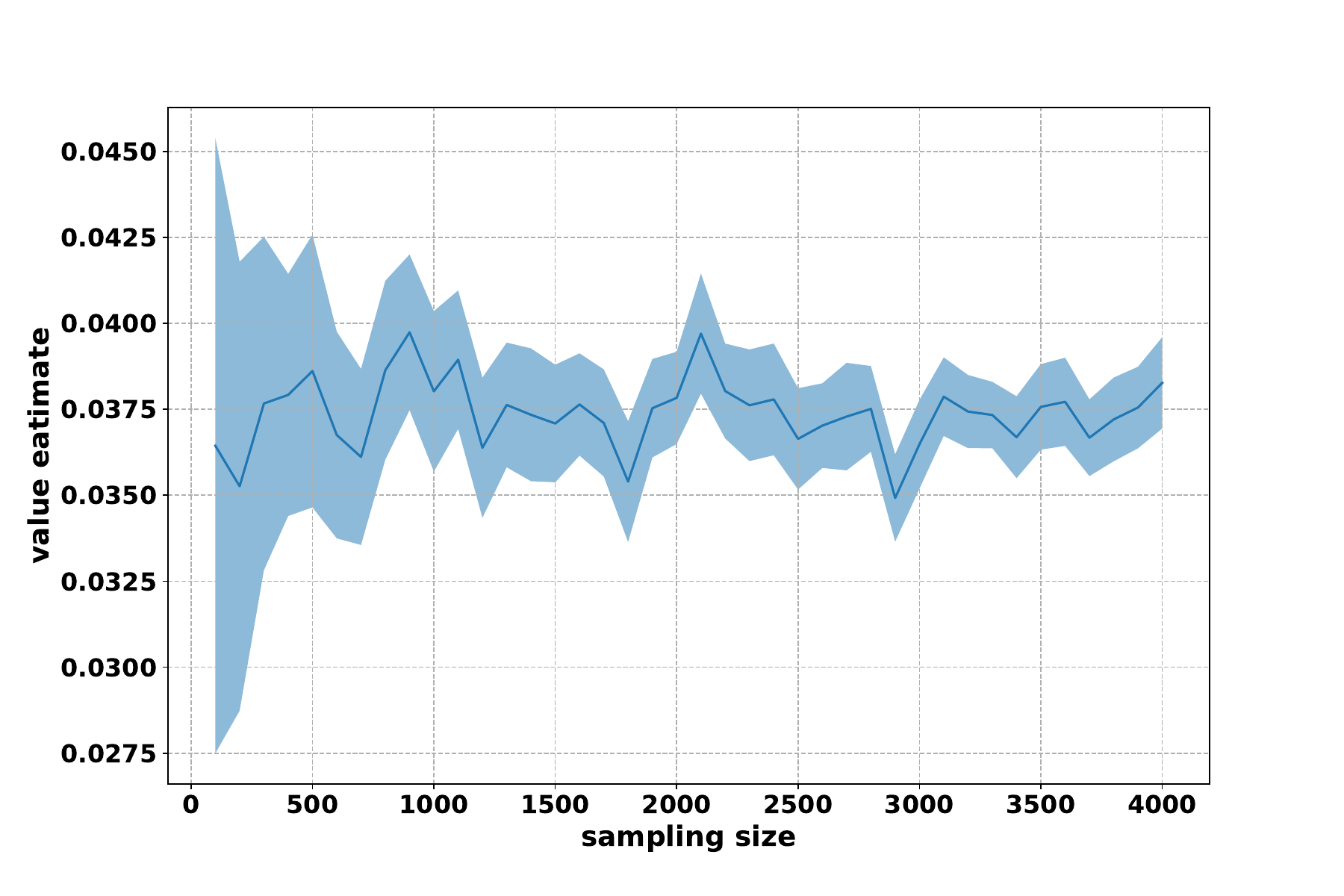}
  \subcaption{}
        \label{f12}
	\end{minipage}
\caption{Estimation value of 20 samples in the FashionMNIST dataset using Algorithm 1, in which the learning algorithm is the Naive Bayes (NB), and the utility function is the test accuracy.
(a) shows the box plot of data Shapley, with a permutation sampling number of 1000 and 50 repetitions of the calculation.
(b) shows the variation of variance of one data point as the sampling number increases. The solid line represents the mean of the estimated values of the data, and the colored area corresponds to the 5th and 95th confidence interval around the mean.
}
\label{FashionMNIST}
\end{figure}

Although the permutation sampling algorithm is easy to operate, sometimes the variance of the data Shapley estimator is too large to be accepted for the data marketplace.
For instance, Figure \ref{f11} shows the box plot of data Shapley of 20 data from the FashionMNIST dataset~\citep{xiao2017fashion}, in which the learning algorithm is the Naive Bayes (NB), the utility function is the test accuracy, the sampling number of permutation is 1000, and the calculation is repeated 50 times.
Figure \ref{f12} is the variation of the variance of one data point as the sampling number increases.
The solid line is the mean of the estimated values of one data repeating 50 times, and the colored area is the 5th and 95th confidence interval around the mean.
These two figures show that even with a large sampling number, the variance is too large compared with the value, which may lead to divergence and failure of transactions in practical application.
Inspired by this, we propose the variance-reduced data Shapley in the following section.

\section{Variance Reduced Data Shapley}~\label{sec:vrds}
This section utilizes stratified sampling to replace the random sampling in robustly estimating data Shapley.

\subsection{Stratified Sampling for Data Shapley}
The stratified sampling approach~\citep{cochran1977sampling} divides the population into several layers according to a specific characteristic and then randomly samples in each layer to form samples~\citep{2000Sampling}.  

\begin{figure}
    \centering
\includegraphics[width=0.8\linewidth]{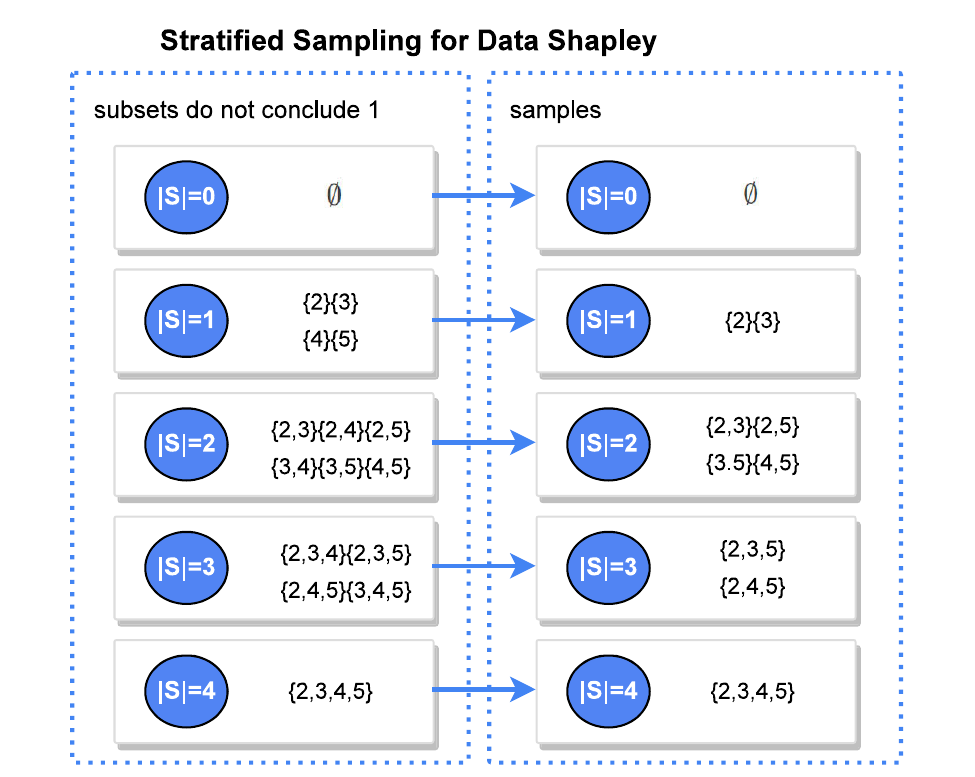}
    \caption{Example of the stratified sampling}
    \label{example}
\end{figure}

For data Shapley, the utility function is usually an evaluation metric such as predictive accuracy, F1, etc. 
The values of evaluation metrics are directly affected by the training sample size.
Therefore, for every data, we consider dividing the coalition sets according to their size when calculating the marginal contribution. Recall $\phi_i(U)$ that can be calculated as
\begin{align}
 \phi_i(U)
 &=\frac{1}{n}\sum_{k=0}^{n-1}\sum_{S\subseteq N\backslash\{i\},|S|=k}\frac{1}{\tbinom{n-1}{k}}(U(S \cup \{i\})-U(S)). 
\end{align}
In words, by grouping the coalitions which do not contain the data $i$, based on their sizes, we have
$n$ strata $S^0, S^1, \dots, S^{n-1}$ such that $S^k=\{S\subseteq N\setminus\{i\},|S|=k\}$ contains all
the coalitions with size $k$. 
For example, suppose we have 5 samples and the index set is $\{1,2,3,4,5\}$. Now we want to calculate the Shapley value of the first datum. We need to know the marginal contribution of the first datum for all data sets that do not contain it. 
To accomplish this, we divide these sets into five categories based on their cardinality, list them separately, and then randomly sample $m_k$ sets from the $k$th stratum to calculate, refer to Figure \ref{example}.
We suppose the
dependency on $U$ when the utility is self-evident and
use $\phi_i$ to represent the value allocated to data $i$.
Let $\phi_{i,k}$ denote the expected marginal contribution of the data $i$ within stratum $S^k$, then
we denote
\begin{equation}
    \phi_{i,k}=\frac{1}{\tbinom{n-1}{k}}\sum_{S\subseteq N\backslash\{i\},|S|=k}(U(S \cup \{i\})-U(S)),
\end{equation}
and it is obvious that data Shapley of data $i$ can be calculated as follows
\begin{equation}
    \phi_i=\frac{1}{n}\sum_{k=0}^{n-1}\phi_{i,k}.
\end{equation}

Suppose that the random variable $X_i(S)=U(S \cup \{i\})-U(S)$ represents the marginal contribution of data $i$ to $S$, and $\phi_ {i,k}$ is its expected marginal contribution to all sets that do not contain $i$ and whose size is $k$.
It is natural to use the sample mean to estimate its expectation to reduce computing costs.
By denoting $m_k$ the number of samples taken from $S^k$, we obtain the estimate of $\phi_i$ as 
\begin{equation}
  \hat{\phi}_i=\frac{1}{n}\sum_{k=0}^{n-1}\hat{\phi}_{i,k}=\frac{1}{n}\sum_{k=0}^{n-1}\frac{1}{m_k}\sum_{j=1}^{m_k}X_{i,k,j},
\end{equation}
where $X_{i,k,j}$ represents the marginal contribution of the $i$th point to the $j$th sample in stratum $S^k$. 
To minimize the variance of $\hat{\phi}_i$, a critical problem is determining the number of samples $m_k$ in each stratum. 
Theorem \ref{variance reduction} provides a feasible solution.

\begin{theorem}\label{variance reduction}
Suppose that using a training set $D$ which contains $n$ i.i.d samples to train a machine learning model and performs on a test set $T$.
A utility function $U:2^N \rightarrow \mathbb{R}$ satisfies the above four axioms.
We utilize the proposed stratified sampling algorithm to estimate the data Shapley (Algorithm 2) with the total number of samples $m$.
To minimize the variance of the data Shapley estimator, the optimal number of samples in the $k$th stratum is given by
\begin{equation}
m_{i,k}^*=m\frac{\sigma_{i,k}}{\sum_{j=0}^{n-1}\sigma_{i,j}}\label{sample proportion},\quad k=0,\dots,n-1,
\end{equation}
where $\sigma_{i,k}$ denotes the 
standard deviation
of $\hat{\phi}_{i,k}$. In addition, it can be proved that the variance obtained by using stratified sampling is no greater than that resulting from permutation sampling.
\end{theorem}

Theorem \ref{variance reduction} shows that the variance obtained by the stratified sampling method is less than or equal to that obtained by the permutation sampling method. Moreover, it reflects that when the total number of samples $m$ is fixed, the number of samples in each layer should be proportional to its standard deviation.
In practical application, we do not know the variance of each layer, so it is necessary to provide an alternative value of variance to determine the sample size. 
Indeed, we can justify that the variance of each layer is bounded by the range of the utility function in that layer. 
It is worth noting that the variance does not increase in machine learning as the number of training samples increases.
Therefore, it is reasonable to assume that the utility range does not increase for the number of layers $k$ (the size of training set of samples in each layer increases for $k$).
Taking full advantage of the above property, Theorem \ref{variance scale} provides a method to allocate the number of samples when the variance is unknown.

\begin{theorem}\label{variance scale}
With the same assumptions of Theorem 1, we further denote that $r_{i,k},i=1,\dots,n,k=0,\dots,n-1$ is the range of set $\{U(S\cup \{i\})-U(S),\text{where }|S|=k\}$. Then there exists a non-increasing function of $f(k)$ that satisfies
\begin{equation}
   \sigma_{i,k}^2 \leq  r_{i,k}^2/4\leq \frac{d_i^2f(k)^2}{16},
\end{equation}
where $d_i$ is a constant related to the $i$th sample.
Then, using the unbounded of $\sigma_{i,k}^2$ to replace its funcnality in \eqref{sample proportion}, we can obtain the approximate $m_k$ as

\begin{equation*}
    \Tilde{m}_k=m\frac{f(k)}{\sum_{j=0}^{n-1}f(j)},\quad k=0,\dots,n-1.
\end{equation*}
\end{theorem}

Theorem \ref{variance scale} proves that the sample size of each stratum can be set as proportional to some non-increasing functions of $k$.
Next, we analyze several specific forms of function $f$.
\begin{itemize}
    \item $f(k)=c$, where $c$ is a constant, then
    $$
    \Tilde{m}_k=m\frac{ c}{\sum_{j=0}^{n-1}c}=\frac{m}{n}.
    $$
    This means that the number of samples is equally distributed to each layer.
    \item $f(k)=\frac{1}{k+1}$, then
    $$
    \Tilde{m}_k=m\frac{ \frac{1}{k+1}}{\sum_{j=0}^{n-1}\frac{1}{j+1}}.
    $$
    Contrary to the above, assuming that $f(k)$ is decreasing with respect to $k$, because the number of samples per layer is also decreasing with respect to the set cardinality.
    \item $f(k)=(k+1)^a, a<0,a\neq -1$, where the absolute value of $a$ reflects the rate at which $f$ changes with $k$.
    In different data sets, different algorithms are used to calculate different amounts of data Shapley. 
    Thus, $a$ can be considered as a tuning parameter that can influence the performance of estimating data Shapley.
    {In machine learning applications, the utility function of a set $S$ is often defined as the loss of the model trained over $S$ for predicting a test point $z$, i.e., $U(S) = l(\mathcal{A}(S),z)$, where  $\mathcal{A}$ represents the underlying learning algorithm that takes in a dataset and outputs a model. Under this utility function definition, the range of $U(S\cup \{i\})-U(S)$---the marginal contribution of any data point $i$ to a subset of size $k$---can be upper bounded by uniform stability of $\mathcal{A}$, defined by $\max_{S\in\mathcal{Z}^k}\max_z\max_i |l(\mathcal{A}(S\cup \{i\}),z) - l(\mathcal{A}(S),z) |$. Prior work~\citep{bousquet2002stability,hardt2016train} have shown that the upper bound of uniform stability of many common learning algorithm at size $k$ is $\mathcal{O}(\frac{1}{k+1})$. These theoretical results shed light on our empirical observation that the best choice of $a$ is usually $-1$, giving rise to the least variance in data Shapley estimation compared to the other possible choices.}
    Therefore, we provide that a suggested interval of $a$ is $[-1,-1/2]$.
    Detailed pieces of evidence will be presented in the experimental section.
\end{itemize}

    

Taking into account that sample size should be an integer, we can set the value of $m_k=\min \{1,\lfloor \Tilde{m}_k \rfloor\}$. However, this implies that additional samples may be left unused as $\sum_{k=0}^{n-1}m_k$ may be lower than $m$.
In this case, we sequentially increase the value of $m_k$ from $k=0$ to $n-1$ until the sum exceed $m$.

So far, we have handled how to use the stratified sampling to reduce the variance of estimated data Shapley. Next, we will give the specific algorithm for the real implementation in Algorithm \ref{stratified sampling data shapley}.


\subsection{VRDS and Its Sample Complexity Analysis}

Algorithm \ref{stratified sampling data shapley} presents the pseudo-code of the stratified sampling algorithm, which first calculates the sampling number of each stratum and then derives the value of each data by taking samples from every stratum.
Note that we refer to the estimated data Shapley by Algorithm \ref{stratified sampling data shapley} as variance reduced data Shapley (VRDS for short).

\begin{algorithm}[!ht]
	\caption{Stratified Sampling Algorithm for VRDS}
	\label{stratified sampling data shapley}
	\begin{algorithmic}[1]
	\REQUIRE
	Training data  $D=\{(x_i,y_i)\}_{i=1}^n$, learning algorithm $\mathcal{A}$, performance score $U$, and sampling number $m$.\\
    \ENSURE Shapley value of training data: $\phi_1,\dots,\phi_n$\\
    \STATE Initialize $\phi_i:=0$ for $i=1,\dots,n$ and $t=0$;\\
    \STATE $m_k:=\min\{1, \lfloor m\frac{f(k)}{\sum_{j=0}^{n-1}f(j)}\rfloor\}$, for $k= 0,\dots,n-1$;\\
    \WHILE{$m-\sum m_k>1$}
    \STATE $m_t:=m_t+1$;
    $t:=t+1$;
    \ENDWHILE
    
	\FOR{$i\in \{1,\dots,n\}$}
	    \FOR{$k\in \{0,\dots,n-1\}$}
	    \STATE 
	    $l:=0$, $\phi_{i,k} :=0$;\\
	    \WHILE{$l \le m_k$}
	    \STATE 
	    $S$:= get a random coalition of $\{1,\dots,i-1,i+1,\dots,n\}$ with size $k$;\\
        \STATE  $\phi_{i,k}:=\phi_{i,k}+(U(S\cup \{i\},\mathcal{A})-U(S,\mathcal{A}))$;\\
       \STATE $l:=l+1;$
	    \ENDWHILE
	    \STATE  $\phi_{i,k}:=\frac{\phi_{i,k}}{m_k}$;\\
	    $\phi_i:=\phi_i+\frac{\phi_{i,k}}{n}$;\\
		\ENDFOR
	  	
	\ENDFOR
	\end{algorithmic}
\end{algorithm}


In this section, we will present sample complexity analysis of VRDS to explain the minimum sampling number in the stratified sampling algorithm to achieve an $(\epsilon,\delta)$-approximation.


\begin{theorem}
Algorithm 2 gets an $(\epsilon,\delta)$-approximation of data Shapley of each data if the sampling number $m$ satisfies
\begin{equation}
        m \ge \max\left(\frac{16\log\frac{2}{\delta}}{17\epsilon^2n^2}\sum_{k=0}^{n-1}\frac{1}{f(k)}\sum_{j=0}^{n-1}f(j),\frac{2\log\frac{2}{\delta}}{\epsilon^2 n^2 (f(n-1))^2}(\sum_{j=0}^{n-1}f(j))^2\right).
\end{equation}
\label{error restriction}
When $f(k)=\frac{1}{k+1}$, it is sufficient to have
\begin{equation}
    m \ge \frac{2\log\frac{2}{\delta}}{\epsilon^2}(\log n +1)^2.
\end{equation}
\end{theorem}

\begin{remark}
From Lemma \ref{permutation number}, we know that the minimum number of samples achieving $(\epsilon,\delta)$- approximation in the permutation sampling algorithm is independent of $n$. 
For the stratified sampling algorithm, since the sampling number depends on stratum number $n$, according to Theorem \ref{error restriction}, it requires the same order (if we choose the suggested tuning parameter $a=-1$) with the permutation sampling up to a log factor about $n$ for $(\epsilon,\delta)$- approximation. 
The stratified sampling algorithm achieves a small variance when the minimum sample size $m$ does not increase significantly with the increase of $n$, which provides a theoretical basis for its wide application.

 
\end{remark}

\section{Experiments}~\label{sec:experiment}
\subsection{Experimental Setting}
\subsubsection{Experimental Objective}

Our experiments have the following three purposes. 
 
First, we show that the stratified sampling algorithm for VRDS can estimate the data Shapley, and the variance of the estimated value by VRDS is smaller than that by the permutation sampling algorithm.
Fortunately, ~\cite{jia2019efficient} propose an elaborate method to compute the exact data Shapley for K-Nearest Neighbor (KNN) algorithm. Therefore, we use KNN in the first experiment. 

Second, the performance of VRDS is directly affected by the tuning parameter $a$, which determines the number of samples per stratum.
In order to explore whether different data sets, sample sizes and algorithms affect the selection of $a$, we use 14 data sets and five commonly used algorithms and conduct experiments with data sizes from 20 to 2000 to obtain the suggested parameter selection scheme of $a$. 

Finally, data quality is a critical indicator for data analysis, mining, and application. 
It is usually necessary to judge the data quality in the data preprocessing stage and remove the data with poor quality~\citep{wang2021unified}.
Data Shapley can be considered as a data quality measurement. 
So one of the applications of data Shapley is removing data with poor quality based on data Shapley for improving the prediction performance.
In this part, we construct a new criterion based on the estimated variance of data Shapley to remove bad quality data and compare it with the results of the permutation sampling algorithm.

\subsubsection{Datasets} 
We evaluate the performance of the stratified sampling algorithm for VRDS on image data and tabular data. 
Table \ref{Data_sets} describes these data sets. 
These data sets are chosen to provide classification problems, with varying dimensionality, and a mixture of problem domains. 
We set data size ranges from 20 to 2000 to verify the algorithm's robustness.

\setlength{\tabcolsep}{0.5mm}{
\begin{table}
    \centering
    \scriptsize
 \begin{tabular}{ p{2cm} p{1.5cm} p{8cm}}
 \hline
Data set
 & Model
&Reference
\\
 \hline
FashionMNIST & LR, KNN & ~\cite{xiao2017fashion}\\
Iris & NB, KNN & ~\cite{fisher1936use}\\
Digits & KNN, Tree &~\cite{alimoglu1996methods}\\
Breast Cancer & SVC, KNN & ~\cite{mangasarian1995breast}
\\
Spam classification & NB, LR  &  \href{Email Spam Classification Dataset CSV|Kaggle}{https://www.kaggle.com/datasets/balaka18/email-spam-classification-dataset-csv}\\
Creditcard & LR & 
~\cite{yeh2009comparisons}\\
Vehicle &LR &~\cite{duarte2004vehicle} \\
Apsfail &LR & \href{ml}{https://archive.ics.uci.edu/ml/datasets/IDA2016Challenge}\\
Phoneme &LR  &\href{Keel}{https://sci2s.ugr.es/keel/dataset.php?cod=105}\\
Wind & LR & \href{wind}{https://www.openml.org/search?type=data\&status=any\&id=847}\\
Pol  & LR & \href{pol}{https://www.openml.org/search?type=data\&status=any\&id=722}\\
Cpu  & LR  &  \href{cpu}{https://www.openml.org/search?type=data\&status=any\&id=796}\\
Fraud &LR  & 
~\cite{dal2015calibrating}\\
2Dplanes &LR  &\href{2dplanes}{https://www.openml.org/search?type=data\&status=active\&id=727} \\

 \hline
\end{tabular}
\caption{Data sets}
\label{Data_sets}
\end{table}
}


\subsubsection{Machine Learning Model and Performance Metric} 
To prove that VRDS is effective for various algorithms, we use Logistic Regression (LR), K-Nearest Neighbor algorithm (KNN), Naive Bayes (NB), Decision Tree (DT), and Support Vector Classification (SVC).
To implement these algorithms, we directly employed the Python package ``scikit-learn''.
To accurately calculate the data Shapley value, another KNN solver in ~\cite{jia2019efficient} is used.
Since the performance metric selection has no direct relationship with the model performance, we always use prediction accuracy as the performance evaluation metric of all algorithms.

We set the sampling number from 100 to 5000 to conduct comparative experiments to test the effect of our VRDS and compare the deviation and variance of the estimated values obtained by VRDS and permutation sampling.
For each experiment, we randomly sampled fifty times for estimation to obtain the variance of the estimated value.

\subsection{Experimental Results}
\begin{figure}
    \centering
\includegraphics[width=.8\linewidth]{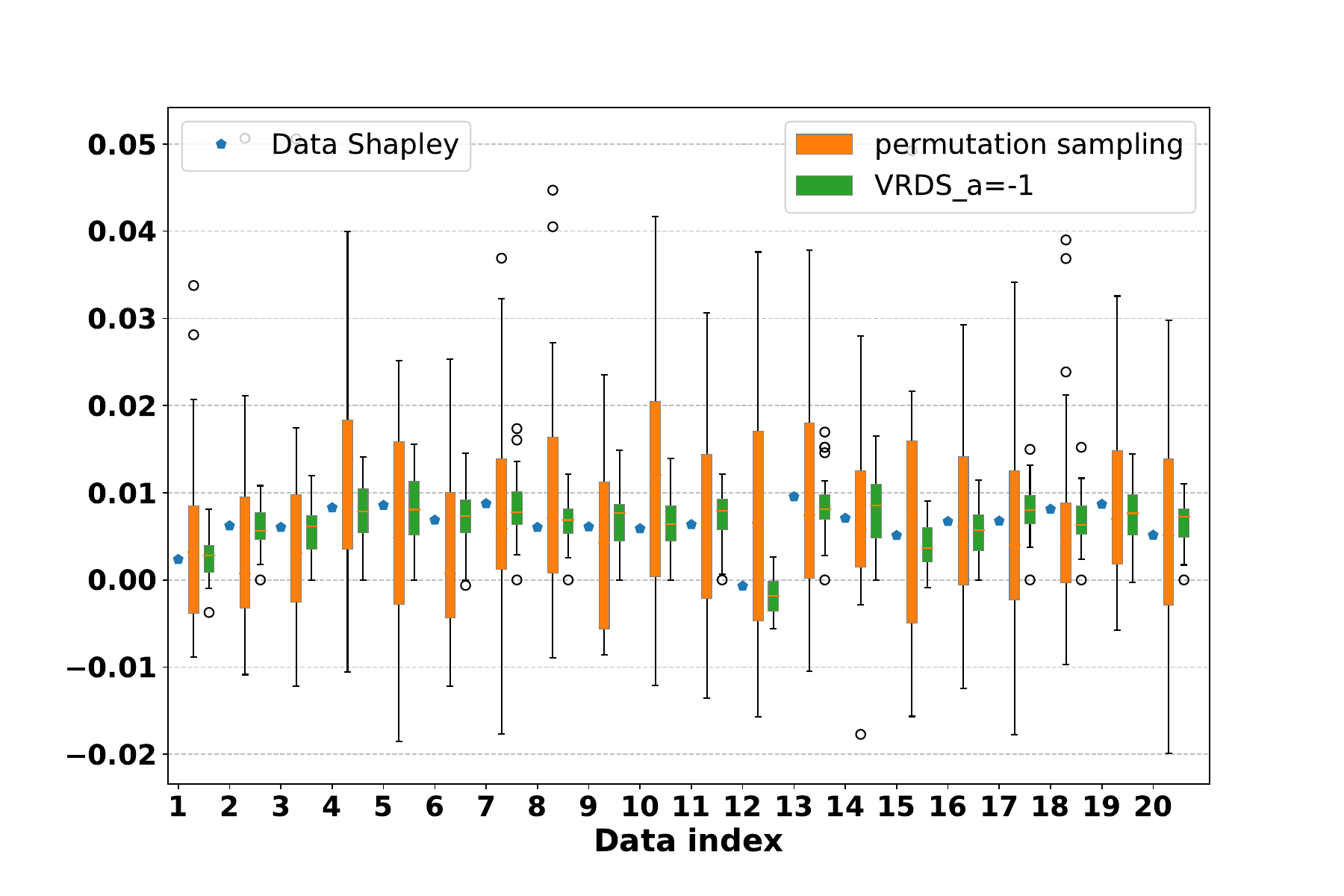}
    \caption{The data Shapley values of 20 data in the FashionMNIST dataset, and their estimates. The learning algorithm is KNN and the utility function is test accuracy. Permutation sampling and $VRDS (a=-1)$ are used, respectively.
    The experiment is repeated 30 times to estimate values.}
    \label{exa_estimated}
\end{figure}

\textbf{Variance Comparison of VRDS.}
We first compare the data Shapley to its estimates obtained by employing two algorithms: permutation sampling and VRDS($a=-1$).
To compute the data Shapley for large datasets, we utilize KNN since ~\cite{jia2019efficient} provides an effective and accurate approach for calculating data Shapley using KNN.
Taking test accuracy as the utility function, we calculated the data Shapley of 100 data points and their estimates using permutation sampling and VRDS$(a=- 1)$.
Figure \ref{exa_estimated} displays the data Shapley and their estimates for 20 of these data points.
Both permutation sampling and VRDS($a=-1$) provide estimates close to exact values, but estimates from VRDS($a=-1$) have much lower variance compared to those from permutation sampling.

\begin{figure}[htbp]
	\centering
	\begin{minipage}{0.49\linewidth}
		\centering     \includegraphics[width=1\linewidth]{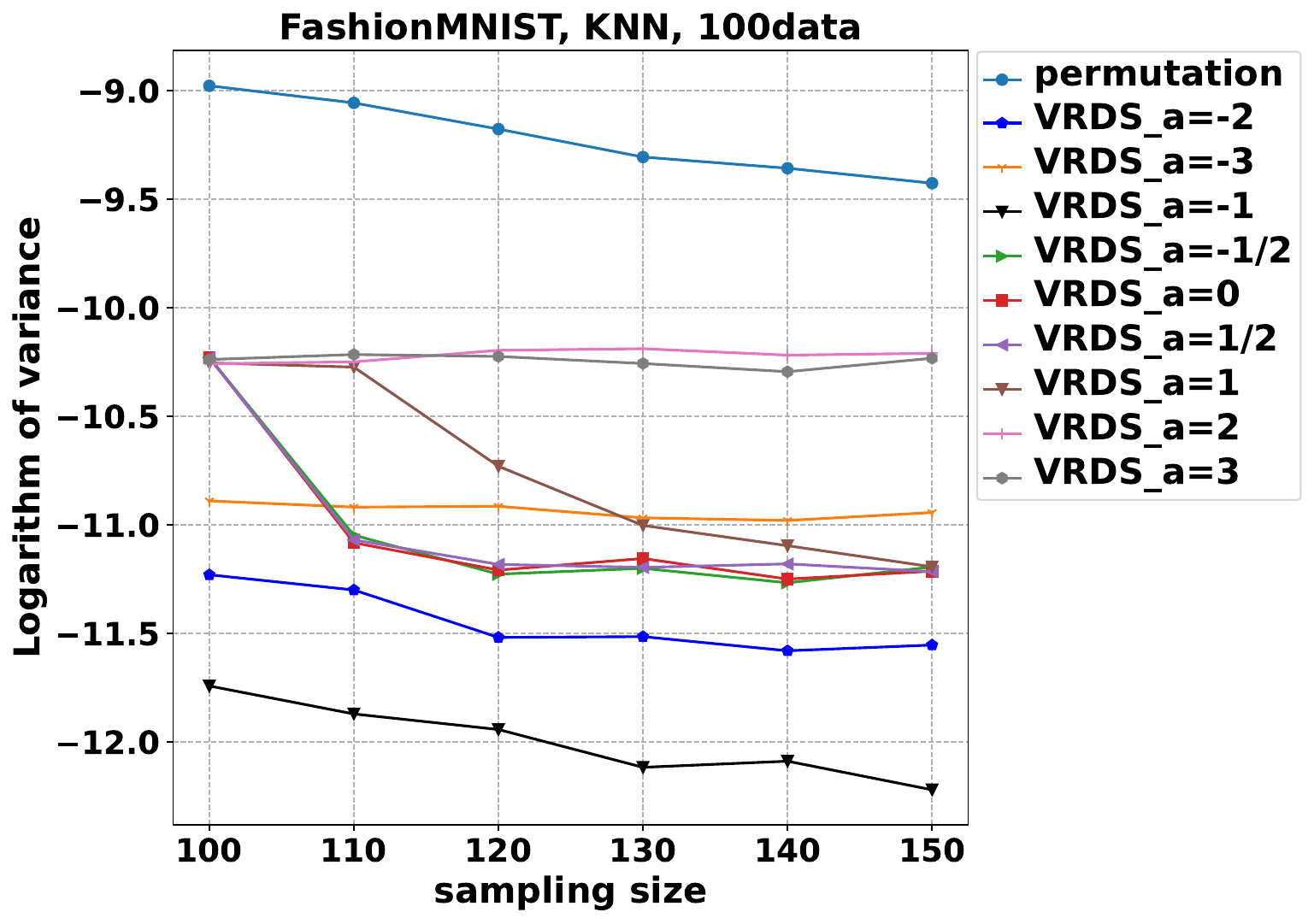}
		\subcaption{}
        \label{es1}
	\end{minipage}
 \begin{minipage}{0.49\linewidth}
		\centering    \includegraphics[width=1\linewidth]{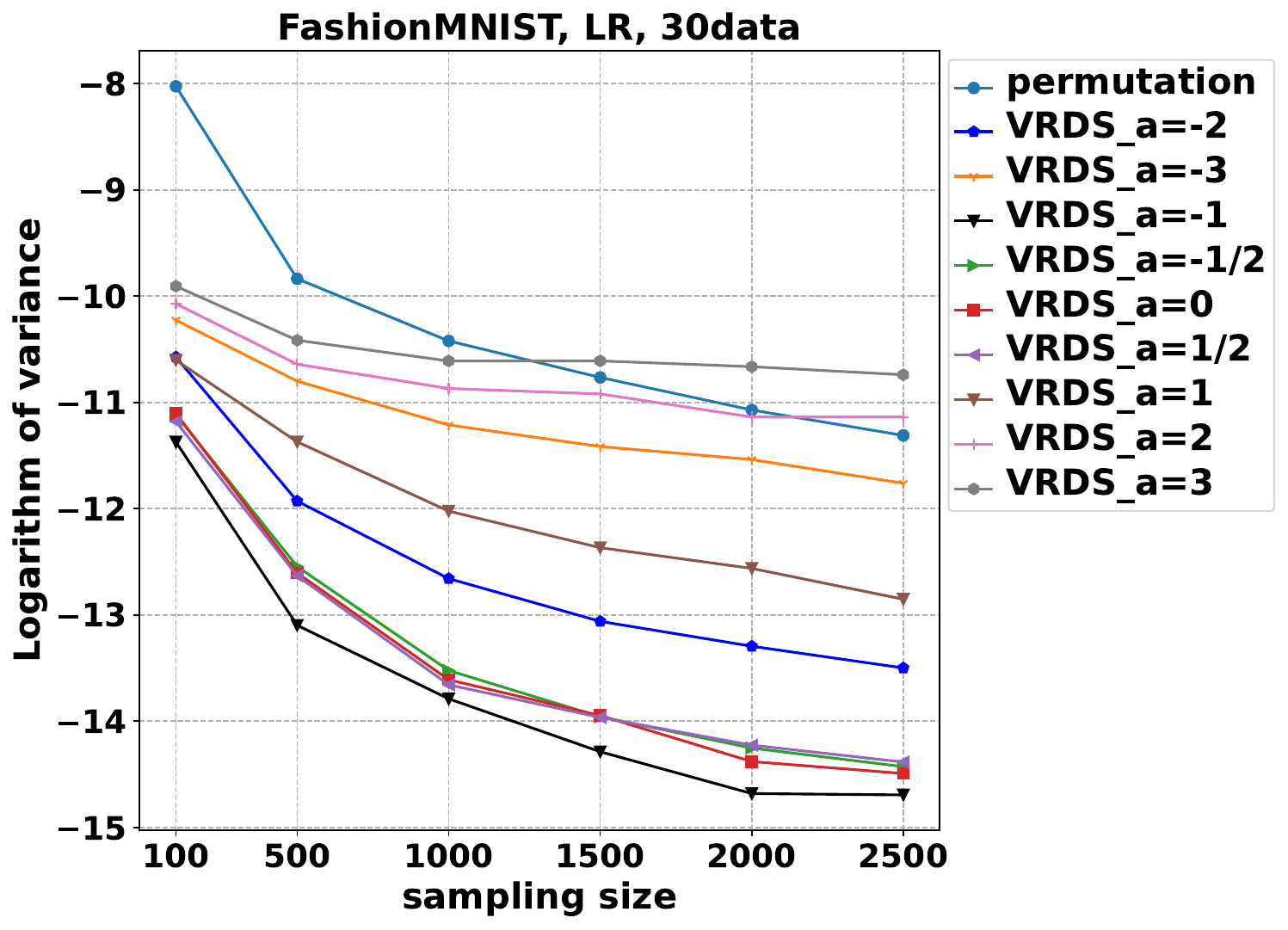}
		\subcaption{}
        \label{es2}
	\end{minipage}
   \caption{The variation of variance with the number of samples when using the permutation sampling algorithm and the VRDS algorithm where the parameter $a$ takes value ranging from -3 to 3. The dataset is FashionMNIST. In Figure (a), the KNN model is used, and the dataset size consists of 100 points. In Figure (b), the LR model is used, and we consider 30 data points.}
 \label{fig_KNN_LR}
\end{figure}

\textbf{Parameter Selection for VRDS.}
We compare the variance reduction effect of VRDS when parameter $a$ takes different values, and different data sets, sample sizes, and algorithms are considered.
We first make intensive attempts on the parameters with the FashionMNIST dataset.
Figure \ref{es1} illustrates the variances of data Shapley estimates for 100 data points computed using the VRDS method with different values of parameter 
$a$, with the KNN machine learning model. 
The horizontal axis denotes the number of samples, ranging from 100 to 150, while the vertical axis represents the logarithm of the corresponding variance.
Figure \ref{es2} presents a scenario where estimates of 30 data Shapley are calculated and the machine learning model is LR.
All estimates are obtained through 30 repetitions.
From the two figures, we can see that in all methods, when the number of samples is fixed, VRDS with $a\in[-1,-1/2]$ has a minor variance (see more experiments in \ref{sec:add_experiment}).
In addition, the variance decreases with the number of samples. 

We use Table \ref{otherdataset} to summarize the variance of the results calculated by different models on different data sets when the algorithm is LR and the number of samples is 150.
It can be seen that the variance of the estimated value obtained by VRDS is obviously smaller than that obtained by the permutation sampling algorithm. 
Moreover, when $a=-1$ and $-\frac{1}{2}$, the VRDS method is better than when $a$ takes other parameters.

\setlength{\tabcolsep}{0.5mm}{
\begin{table}
    \centering
 \begin{tabular}{ p{2cm} p{2cm} p{2cm}   p{2cm} p{2cm} p{2cm}}
 \hline
Dataset
 & Permutation
&$a=0$
&$a=-1/2$
&$a=-1$
&$a=-2$
\\
 \hline
 Cpu   & 70.36   &4.05&   2.18&\textbf{1.97}&2.50\\
 Pol  & 59.15   &2.36&1.76&\textbf{1.73}&2.39\\
Vehicle & 58.22  &2.60&2.63&\textbf{2.37}&3.09\\
2dplanes & 59.79   &2.42&2.41&\textbf{2.15}&3.02\\
Creditcard&55.42 &2.31&\textbf{2.27}&2.55&3.19\\
Apsfail &76.24  &3.86&2.21&\textbf{1.77}&2.27\\
Phoneme &62.29  &2.97&2.59&\textbf{2.53}&3.18\\
Fraud &78.83  &4.92&2.42&\textbf{1.84}&2.76\\
Wind & 72.05  &4.22&2.21&\textbf{1.81}&2.39\\
 \hline
\end{tabular}
\caption{A summary of the variance of data Shapley estimator on different datasets when parameter $a$ takes different values in the VRDS algorithm, with LR as the base model. The data volume is 100, the sampling number is 150, and the unit of variance is $10^{-6}$. The best result is highlighted in bold. All estimates are obtained through 30 repetitions.}
\label{otherdataset}
\end{table}
}

\begin{figure}
    \centering
    \includegraphics[width=1\linewidth]{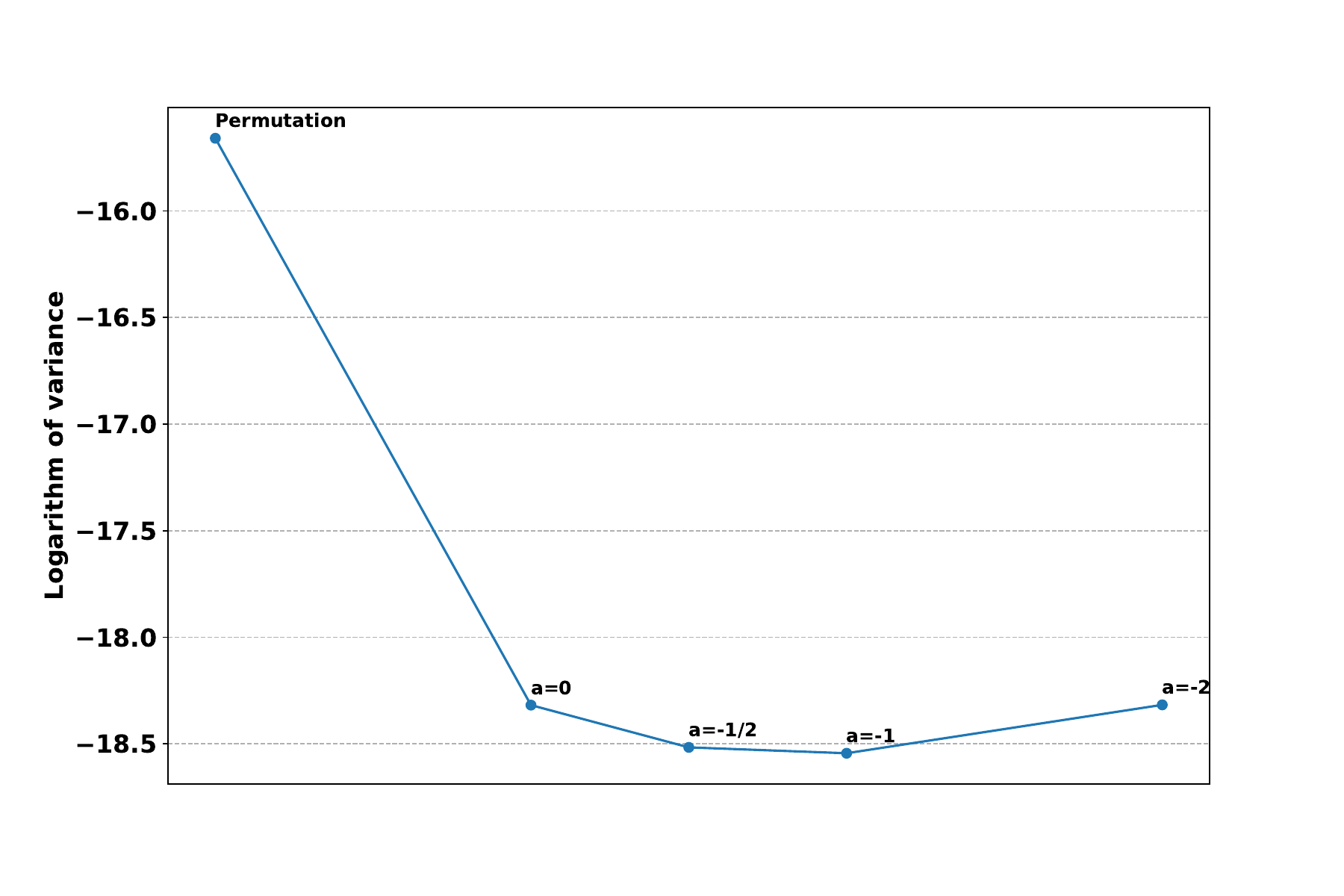}
    \caption{Estimation of 2000 data Shapley in FashionMNIST data set, wherein the optimization algorithm is LR and the number of samples is 2200.}
    \label{fm_2000}
\end{figure}

We also examine whether VRDS is also effective when the data set is large. 
Figure \ref{fm_2000} shows data Shapley estimated by VRDS of 2000 points in the FashionMNIST data set.
The algorithm is LR, and the number of samples is 2200.
We can see that the variance of the results calculated by Permutation sampling is the largest, and the variance is the smallest when $a=-1$.

\textbf{Data Group Removal.}
We evaluate the VRDS by comparing the performance on the data group removal task.
The so-called data group means that a data source contains multiple data.
In fact, in a real data transaction scenario, a data provider usually provides a data set, and the data sets of providers jointly form a training set for a machine learning task.
So in this experiment, we consider the value of the data group.
The data group removal task is to remove groups of data to understand their impact on a model.
A better data value estimation algorithm can better identify the importance of data. 
Therefore, when the data group with the highest (lowest) value estimation is removed, a better data value estimation method will result in faster (slower) performance degradation.

\begin{figure}[htbp]
	\centering
	\begin{minipage}{0.49\linewidth}
		\centering
		\includegraphics[width=1\linewidth]{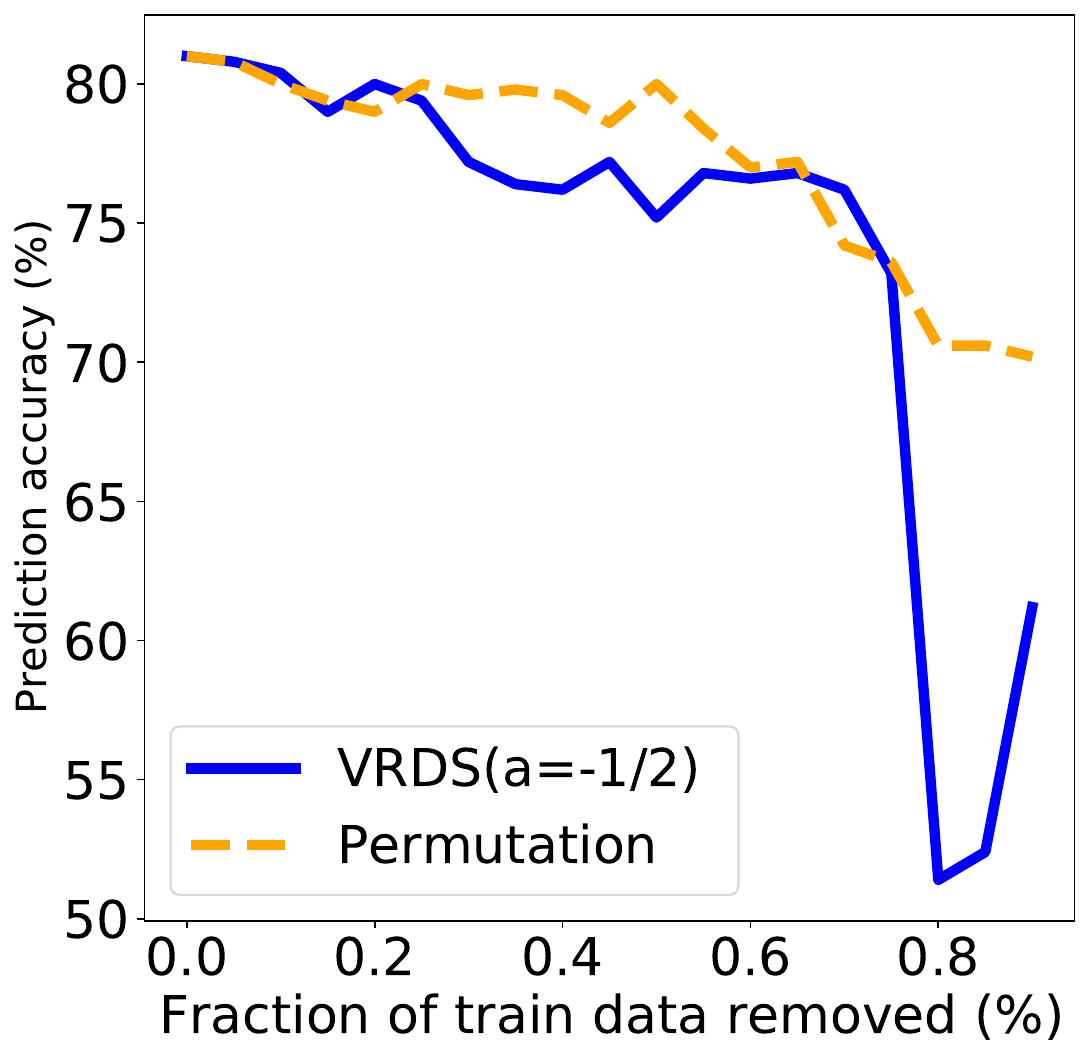}
		\subcaption{}
        \label{dr1}
	\end{minipage}
	\begin{minipage}{0.49\linewidth}
		\centering
		\includegraphics[width=1\linewidth]{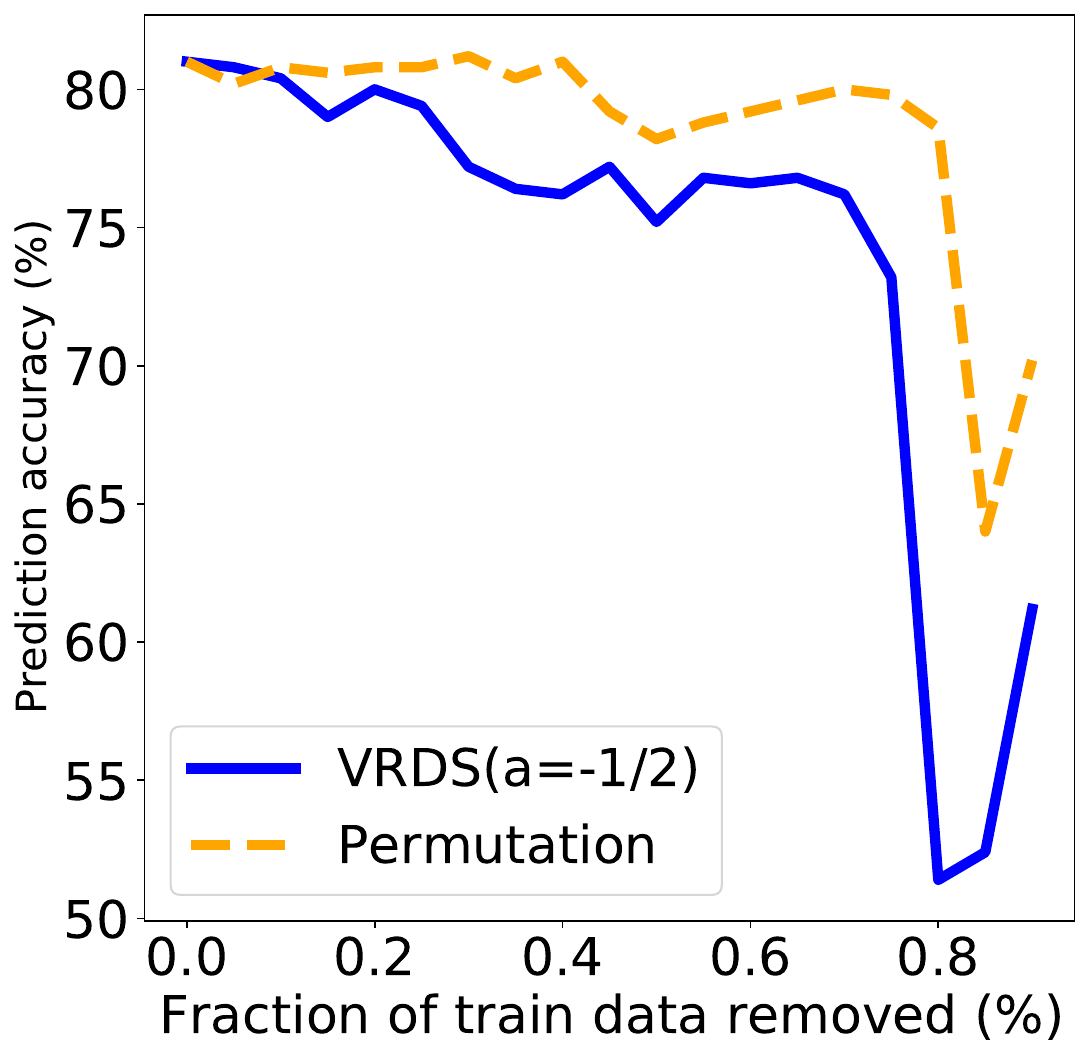}
        \subcaption{}
        \label{dr2}
	\end{minipage}
\caption{Data group removal experiment on the FashionMNIST data set. There are $20$ groups and each group has $100 $ data. VRDS ($a = - \frac{1}{2}$) and permutation sampling algorithms are used to calculate the value of data groups.
 Figure \ref{dr1} shows the removal from the most valuable data group. Figure \ref{dr2} shows the removal from large to small according to the values of $estimated\ data\ Shapley-100 \times variance$. }
\label{data_removal}
\end{figure}

Figure \ref{data_removal} shows the data group removal experiment on the FashionMNIST dataset.
The data has been randomly partitioned into 20 groups, and each group consists of 100 data points. Notably, in practical applications, the dataset of each group generally derives from a single company.

VRDS with $a = - \frac{1}{2}$ and permutation sampling algorithms are used to calculate the value of data groups.
Figure \ref{dr1} shows the removal from the most valuable data group. It can be seen that the performance of VRDS with $a = - \frac{1}{2}$ decreases faster than that of permutation sampling. 
Figure \ref{dr2} shows the removal according to the value of $estimated\ data\ Shapley-100 \times variance$, the prediction accuracy of the VRDS with $a = - \frac{1}{2}$ algorithm decreases faster.
This indicates that VRDS can identify important data groups more accurately by simultaneously combining the estimated data Shapley and its variance.
 
So far, we have demonstrated the effectiveness of VRDS in approximating data Shapley and obtained the suggested parameter $a$ that minimizes the variance of the estimated value through various experiments on a large number of different data sets. 
Finally, the data group removal experiment shows that combining the data value and variance can quickly identify important data sets.
To show more evidence of the results, we further present more experiments in \ref{sec:add_experiment}.

\section{Conclusion}\label{sec:conclusion}

In this work, we propose a more robust data Shapley estimation method, VRDS, which is based on a stratified sampling technique and can obtain data Shapley estimates with minor variance.
We obtain the optimal sampling number of each stratum in the stratified sampling and provide the sample complexity analysis.
Experiments on various data sets show that the variance of estimates calculated by VRDS is always smaller than that calculated by the permutation sampling algorithm.
Furthermore, we also get the suggested parameter interval that can obtain the estimation value with minimum variance on these data sets. 
Finally, the application of VRDS in data quality identification is also discussed. We find that simultaneously considering value and variance can recognize the most valuable data faster.
For future work, we wish to continue applying the stratified sampling algorithm to other data Shapley estimation methods~\citep{jia2019towards,jia2019efficient,kwon2022beta,ghorbani2020distributional} to further reduce the variance of data Shapley estimates to promote the development of the data marketplace.


\section*{Acknowledgements}~\label{sec:ack}
We would like to thank Jiachen T. Wang (Ph.D. student at Princeton University). He provided insightful comments for the manuscript and helped us discuss the implementation.



\appendix
\section*{Appendix}\label{sec:appendix}

\renewcommand{\thesubsection}{Appendix~\Alph{subsection}}

\subsection{Proof of Lemma \ref{permutation number}}
\setcounter{equation}{0}
\renewcommand{\theequation}{A.\arabic{equation}}
Given $r_i$, the range of marginal contributions of data $i$, and $r$, the range of marginal contributions of all data, according to Hoeffding's inequality, we obtain 
\begin{equation}
    P[|\hat{\phi}_i-\phi_i|\ge \epsilon]\leq 2 \exp {\left(-\frac{2m\epsilon^2}{r_i^2}\right)}.
\end{equation}
To get an $(\epsilon,\delta)$- approximation, we should bound the right band side by $\delta$, which means
$2 \exp {(-\frac{2m\epsilon^2}{r_i^2})}\le \delta$, then we have
\begin{equation}
    m \ge  \frac{\log (2/\delta)r_i^2}{2\epsilon^2}.
\end{equation}
In particular, if the utility function is the prediction accuracy, we have $0\le U\le 1$. 
Therefore, the range of the marginal contributions $-1 \le r \le 1$.
Since $ r_i \le r$, $i=1,\dots,n$ and $r^2\le 1$, it is sufficient to make 
\begin{equation}
    m \ge  \frac{\log (2/\delta)}{2\epsilon^2}.
\end{equation}

\subsection{Proof of Theorem \ref{variance reduction}}
\setcounter{equation}{0}
\renewcommand{\theequation}{B.\arabic{equation}}
We divide the proof into two steps. 
The first step is to calculate the sampling number in each layer to minimize the variance. 
The second step is to prove that the variance of the results obtained by the stratified sampling is not greater than that obtained by the permutation sampling.

(1) Firstly, optimize the sampling number per layer.

The variance of data $i$ using stratified sampling is
\begin{equation}
Var(\hat{\phi}_i)=Var\left(\frac{1}{n}\sum_{k=0}^{n-1}\hat{\phi}_{i,k}\right)=\frac{1}{n^2}\sum_{k=0}^{n-1}Var(\hat{\phi}_{i,k})=\frac{1}{n^2}\sum_{k=0}^{n-1} \frac{\sigma_{i,k}^2}{m_{i,k}},
\end{equation}
due to the independent property of samples in each layer.

Our goal is to determine $m_{i,0}, m_{i,1},\dots,m_{i,n-1}$ such that
\begin{align}
    &\min\quad Var(\hat{\phi}_i)=\frac{1}{n^2}\sum_{k=0}^{n-1} \frac{\sigma_{i,k}^2}{m_{i,k}},\\
    &\quad\text{s.t.}\quad m=\sum_{k=0}^{n-1}m_{i,k}.
\end{align}

Using the Lagrange multiplier method, we get 
\begin{equation}
    L=Var(\hat{\phi}_i)+\lambda\left(m-\sum_{k=0}^{n-1}m_{i,k}\right)=\frac{1}{n^2}\sum_{k=0}^{n-1} \frac{\sigma_{i,k}^2}{m_{i,k}}+\lambda\left(m-\sum_{k=0}^{n-1}m_{i,k}\right).
\end{equation}
Take the partial derivatives with respect to $m_{i,k}$ and $\lambda$,
\begin{align*}
\frac{\partial L}{\partial m_{i,k}}&=-\frac{1}{n^2}\frac{\sigma_{i,k}^2}{m_{i,k}^2}-\lambda=0,\\
\frac{\partial L}{\partial \lambda}&=m-\sum_{k=0}^{n-1}m_{i,k}=0,
\end{align*}
we get
\begin{equation}
m_{i,k}^*=m\frac{\sigma_{i,k}}{\sum_{j=0}^{n-1}\sigma_{i,j}}.
\end{equation}

(2) Then, we prove that the variance of the estimator using the stratified sampling is less than or equal to that using the permutation sampling.

For a random variable $X$ (assuming that it is fixed to a certain point $i$, for convenience, the subscript $i$ does not appear in the following proof), we estimate $\phi$ as the mean of $m$ random samples $x_1,x_2,\dots,x_m$, taken from the population of marginal contributions of the player. 
Denote estimator adopting the permutation sampling as $\hat{\phi}_p$, that is  $\hat{\phi}_p=\frac{1}{m}\sum_{i=1}^{m}x_i$. It is an unbiased estimation of $\phi$, and variance of $\hat{\phi}_p$ is 
\begin{align}
Var(\hat{\phi}_p)&=\frac{1}{m} Var(X)\\
&=\frac{1}{m}\{E[Var(X|Y)]+Var[E(X|Y)]\}\\
&=\frac{1}{m}\sum_{k=0}^{n-1}\frac{1}{n}\sigma_k^2+\frac{1}{m}Var[E(X|Y)],
\end{align}
where the second equality is based on the law of total variance, and we further assume that $Y=\{S^0, S^1, \dots, S^{n-1}\}$ represents the sample stratum.

Due to 
\begin{equation}
 \frac{1}{m}Var[E(X|Y)]=\frac{1}{m}[\sum_{k=0}^{n-1}\frac{1}{n}(\hat{\phi}_{k}-\phi_{k})^2]\geq 0, 
\end{equation}
we have
\begin{equation}
Var(\hat{\phi}_p)\geq \frac{1}{m}\sum_{i=0}^{n-1}\frac{1}{n}\sigma_k^2.\label{random variance}
\end{equation}
According to Eq.(\ref{sample proportion}), we obtain that the variance of estimator using the stratified sampling $Var(\hat{\phi}_t)$ is
\begin{align}
Var(\hat{\phi}_s)&=\frac{1}{n^2}\sum_{k=0}^{n-1}{\frac{\sigma_k^2}{m_k}}\\
&=\frac{1}{m}\left(\sum_{k=0}^{n-1}\frac{\sigma_k}{n}\right)^2\\
&\leq \frac{1}{m}\sum_{k=0}^{n-1}\frac{1}{n}\sigma_k^2.
\label{variance of stratified}
\end{align}
The first equality follows from Eq.(\ref{sample proportion}) and the second one is due to Cauchy Swarchz inequality. That is 
\begin{equation*}
    \left(\sum_{k=0}^{n-1}\frac{\sigma_k}{n}\right)^2 \leq \left(\sum_{k=0}^{n-1}\frac{1}{n^2}\right)\left(\sum_{k=0}^{n-1}\sigma_k^2\right)\leq \frac{1}{n}\sum_{k=0}^{n-1}\sigma_k^2.
\end{equation*}
Combining the above inequality Eq.(\ref{random variance}) and Eq.(\ref{variance of stratified}) proves the theorem.

\subsection{Proof of Theorem \ref{variance scale}}
\setcounter{equation}{0}
\renewcommand{\theequation}{C.\arabic{equation}}
Let's consider one data first, assuming that the index of this data is $i$.
Let $\Delta_{i,k}^{max}$ denote the maximum value  of marginal contribution of data $i$ in stratum $k$, that is $\Delta_{i,k}^{max}=\max_{S\subset N \backslash \{i\},|S|=k}\{U(S\cup \{i\})-U(S)\}.$
Similarly, the minimum value is defined as $\Delta_{i,k}^{min}=\min_{S\subset N \backslash \{i\},|S|=k}\{U(S\cup \{i\})-U(S)\}.$
Denote $r_{i,k}$ as the range of  $U(S\cup \{i\})-U(S)$ in stratum $k$, $k=0,1,\dots,n-1$.
Now, let us observe that for any random variable bounded between two values ($\Delta_{i,k}^{max}$ and $\Delta_{i,k}^{min}$ in this case), the maximum variance is reached when this variable takes the two extreme values with the same probability $\frac{1}{2}$~\citep{castro2009polynomial}. 
Thus we have
\begin{align}
    \sigma_{i,k}^2 \le &\frac{1}{2}\left(\Delta_{i,k}^{max}-\frac{\Delta_{i,k}^{max}+\Delta_{i,k}^{min}}{2}\right)^2+\frac{1}{2}\left(\Delta_{i,k}^{min}-\frac{\Delta_{i,k}^{max}+\Delta_{i,k}^{min}}{2}\right)^2\\
    &=\frac{(\Delta_{i,k}^{max}-\Delta_{i,k}^{min})^2}{4}\\
    &=\frac{r_{i,k}^2}{4}.
    \label{variance and range}
\end{align}

On the other hand, suppose $f(k)$ is a  non-increasing function of $k$, since $n$ is finite, we can always find $b_i$ and $d_i$, subject to  $b_i=\min_{k}\frac{r_{i,k}}{f(k)}$, and $d_i=\max_{k}\frac{r_{i,k}}{f(k)}$. 
Then we get $b_if(k)\le r_{i,k} \le d_if(k)$, $k=0,1,\dots,n-1$.

Combining Eq.(\ref{variance and range}), we can give the upper
bounds on the variance $\sigma_{i,k}^2$,
\begin{equation}
    \sigma_{i,k}^2\leq \frac{d_i^2f(k)^2}{4}.
\end{equation}
Substitute $\frac{d_if(k)}{2}$ for $\sigma_{i,k}$ in Eq.(\ref{sample proportion}), 
thus we obtain 
\begin{equation}
\Tilde{m}_k=m\frac{f(k)}{\sum_{j=0}^{n-1}f(j)}.
\end{equation}
It is suitable for all data.

\subsection{Proof of Theorem \ref{error restriction}}
\setcounter{equation}{0}
\renewcommand{\theequation}{D.\arabic{equation}}
To prove Theorem \ref{error restriction}, we first provide the following lemma that has been shown in Theorem 4.2 of~\cite{2021Approximating}. We rewrite this lemma by using the notations of this paper. 
\begin{lemma}
\label{the4.2}
 Let $\hat{\phi}_i$ be the estimator of data Shapley $\phi_i$ adopting Algorithm \ref{stratified sampling data shapley}.
 Denote the mean and variance of $k$th stratum of data $i$ as $u_{i,k}$ and $\sigma_{i,k}^2$, respectively. 
 Let ${X_{i,k,j}},j=0,\dots,m_k$ be independent random variables which denote the marginal contribution of data $i$ in stratum $k$, so
 $-1 \le {X_{i,k,j}} \le 1,j=0,\dots,m_k,$ 
 $\chi_{i,k}=\frac{1}{m_k}\sum_{j=0}^{m_k}X_{i,k,j}$ is their average and $\hat{\phi}_i=\frac{1}{n}\sum_{k=0}^{n-1}\chi_{i,k}$. Then:
 \begin{align*}
    \mathbb{P}\left (|\hat{\phi_i}-\phi_i|\ge \sqrt{4\log (2/t)\sum_{k=0}^{n-1}\left(\frac{1}{17m_k}+\frac{\sigma_{i,k}^2}{2m_k}\right )\frac{1}{n^2}}\right)\le t.
\end{align*}
\end{lemma}

\textbf{Proof of Theorem \ref{error restriction}} 
\setcounter{equation}{0}
\renewcommand{\theequation}{D.\arabic{equation}}
According to Lemma \ref{the4.2}, we obtain
\begin{align*}
    \mathbb{P}(|\hat{\phi_i}-\phi_i|\ge \epsilon)\le 2 \exp \left(-\frac{\epsilon^2}{4\sum_{k=0}^{n-1}(\frac{1}{17m_k}+\frac{\sigma_{i,k}^2}{2m_k})\frac{1}{n^2}}\right).
\end{align*}
Since $\sigma_{i,k}^2\le \frac{d_i^2f(k)^2}{4} $ and $m_k \ge \frac{\Tilde{m}_k}{2}$, we can bound it as 
\begin{align*}
    & 2 \exp \left(-\frac{\epsilon^2}{4\sum_{k=0}^{n-1}\frac{1}{17m_k}\frac{1}{n^2}+4\sum_{k=0}^{n-1}\frac{\sigma_{i,k}^2}{2m_k}\frac{1}{n^2}}\right)\\
    & \le 2 \exp \left(-\frac{\epsilon^2}{4\sum_{k=0}^{n-1}\frac{2}{17m f(k)}\sum_{j=0}^{n-1}f(j)\frac{1}{n^2}+4\sum_{k=0}^{n-1}\frac{d_i^2f(k)^2}{4m f(k)}\sum_{j=0}^{n-1}f(j)\frac{1}{n^2}}\right)\\
    & \le 2 \exp \left(-\frac{\epsilon^2}{\frac{8}{17m n^2}\sum_{k=0}^{n-1}\frac{1}{f(k)}\sum_{j=0}^{n-1}f(j)+\frac{d_i^2}{mn^2}(\sum_{j=0}^{n-1}f(j))^2}\right)\\
    & \le \max \left(2 \exp\left(-\frac{\epsilon^2}{\frac{16}{17m n^2}\sum_{k=0}^{n-1}\frac{1}{f(k)}\sum_{j=0}^{n-1}f(j)}\right) , 2 \exp \left(-\frac{\epsilon^2}{\frac{2d_i^2}{m n^2}(\sum_{j=0}^{n-1}f(j))^2}\right)\right).
\end{align*}
Setting $2 \exp\left(-\frac{\epsilon^2}{\frac{16}{17m n^2}\sum_{k=0}^{n-1}\frac{1}{f(k)}\sum_{j=0}^{n-1}f(j)}\right)\le \delta$ and $2 \exp \left(-\frac{\epsilon^2}{\frac{2d_i^2}{m n^2}(\sum_{j=0}^{n-1}f(j))^2}\right)\le \delta $ yields
\begin{align}
    m \ge \frac{16\log\frac{2}{\delta}}{17\epsilon^2n^2}\sum_{k=0}^{n-1}\frac{1}{f(k)}\sum_{j=0}^{n-1}f(j),
\end{align}
and
\begin{align}
    m\ge \frac{2d_i^2\log\frac{2}{\delta}}{\epsilon^2n^2}(\sum_{j=0}^{n-1}f(j))^2.
    \label{d_i}
\end{align}
So
\begin{align*}
    m \ge \max\left(\frac{16\log\frac{2}{\delta}}{17\epsilon^2n^2}\sum_{k=0}^{n-1}\frac{1}{f(k)}\sum_{j=0}^{n-1}f(j),\frac{2d_i^2\log\frac{2}{\delta}}{\epsilon^2n^2}(\sum_{j=0}^{n-1}f(j))^2\right)
\end{align*}
Since $d_i=\max_{k}\frac{r_{i,k}}{f(k)}\le\frac{r_i}{f(n-1)}$ and $r_i\le 1$, Eq.(\ref{d_i}) can be bound as
\begin{align*}
     m\ge \frac{2\log\frac{2}{\delta}}{\epsilon^2 n^2 (f(n-1))^2}(\sum_{j=0}^{n-1}f(j))^2.   
\end{align*}
Therefore,
\begin{align*}
    m \ge \max\left(\frac{16\log\frac{2}{\delta}}{17\epsilon^2n^2}\sum_{k=0}^{n-1}\frac{1}{f(k)}\sum_{j=0}^{n-1}f(j),\frac{2\log\frac{2}{\delta}}{\epsilon^2 n^2 (f(n-1))^2}(\sum_{j=0}^{n-1}f(j))^2\right).
\end{align*}
When $f(k)=\frac{1}{k+1}$,
since 
\begin{align*}
    \sum_{j=0}^{n-1}\frac{1}{j+1}\le \log n+1
\end{align*}
and $\sum_{k=0}^{n-1}(k+1)=\frac{n(n+1)}{2}$,
we sufficiently have
\begin{align*}
    m \ge \max\left(\frac{8\log\frac{2}{\delta}}{17\epsilon^2}\frac{(n+1)(\log n+1)}{n},\frac{2\log\frac{2}{\delta}}{\epsilon^2}(\log n +1)^2\right).   
\end{align*}
Due to $\frac{2\log\frac{2}{\delta}}{\epsilon^2}(\log n +1)^2 \ge \frac{8\log\frac{2}{\delta}}{17\epsilon^2}\frac{(n+1)(\log n+1)}{n}$ for large $n$,
it is sufficient to let
\begin{align*}
    m \ge \frac{2\log\frac{2}{\delta}}{\epsilon^2}(\log n +1)^2.    
\end{align*}

 \subsection{Experiment Additional Results}~\label{sec:add_experiment}
\begin{figure}[htbp]
	\centering
	\begin{minipage}{0.24\linewidth}
		\centering
		\includegraphics[width=1\linewidth]{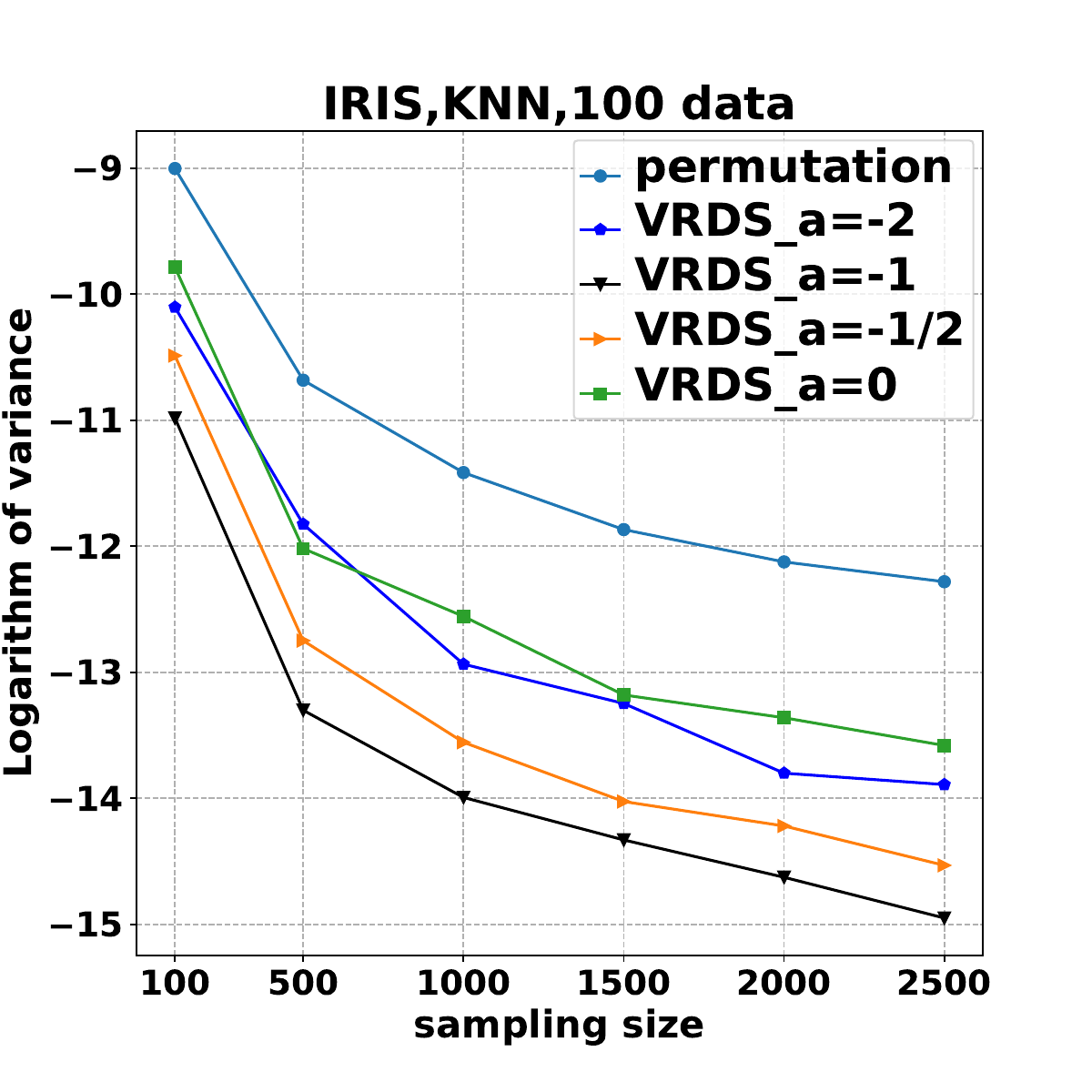}
	\end{minipage}
	\begin{minipage}{0.24\linewidth}
		\centering
		\includegraphics[width=1\linewidth]{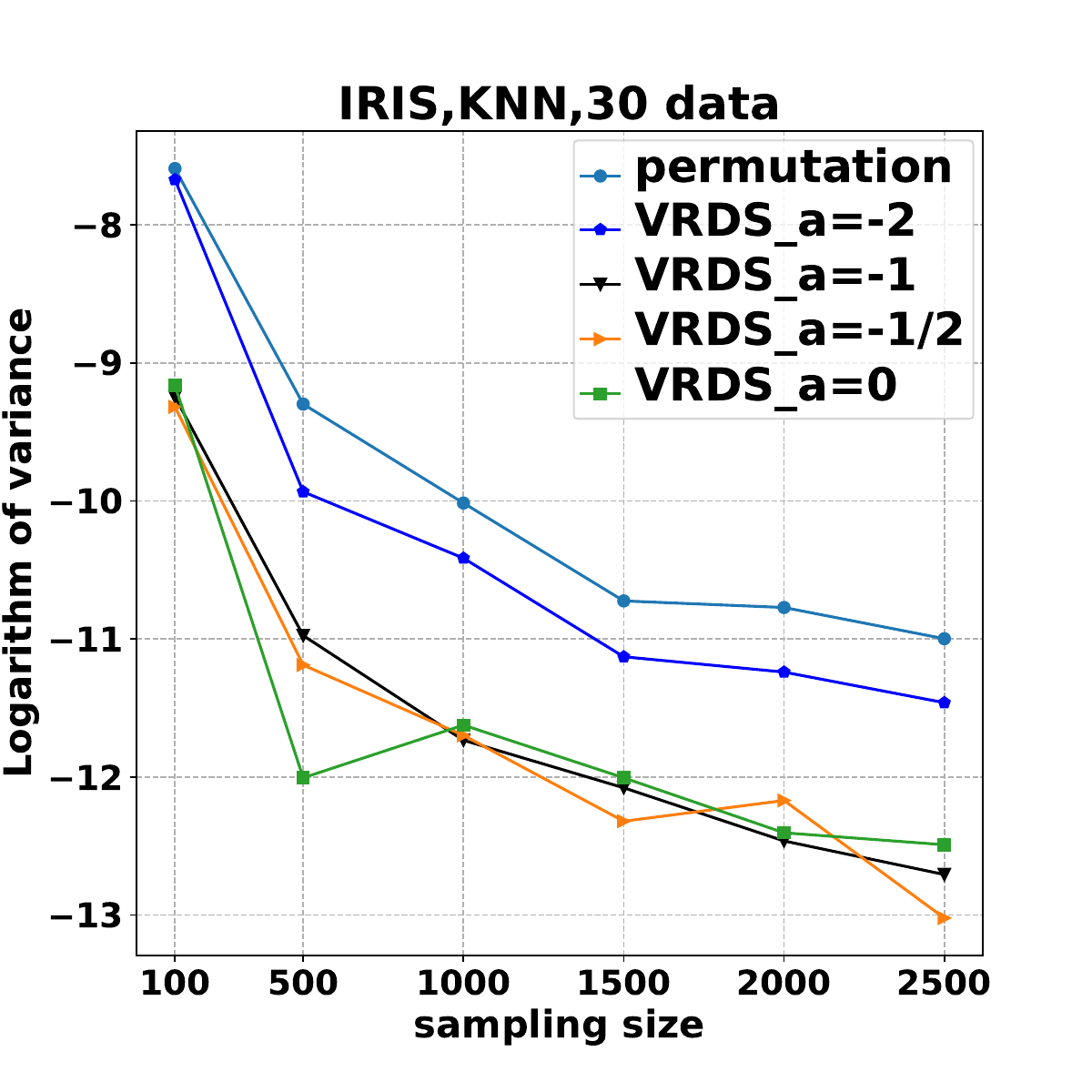}
	\end{minipage}
	\begin{minipage}{0.24\linewidth}
		\centering
		\includegraphics[width=1\linewidth]{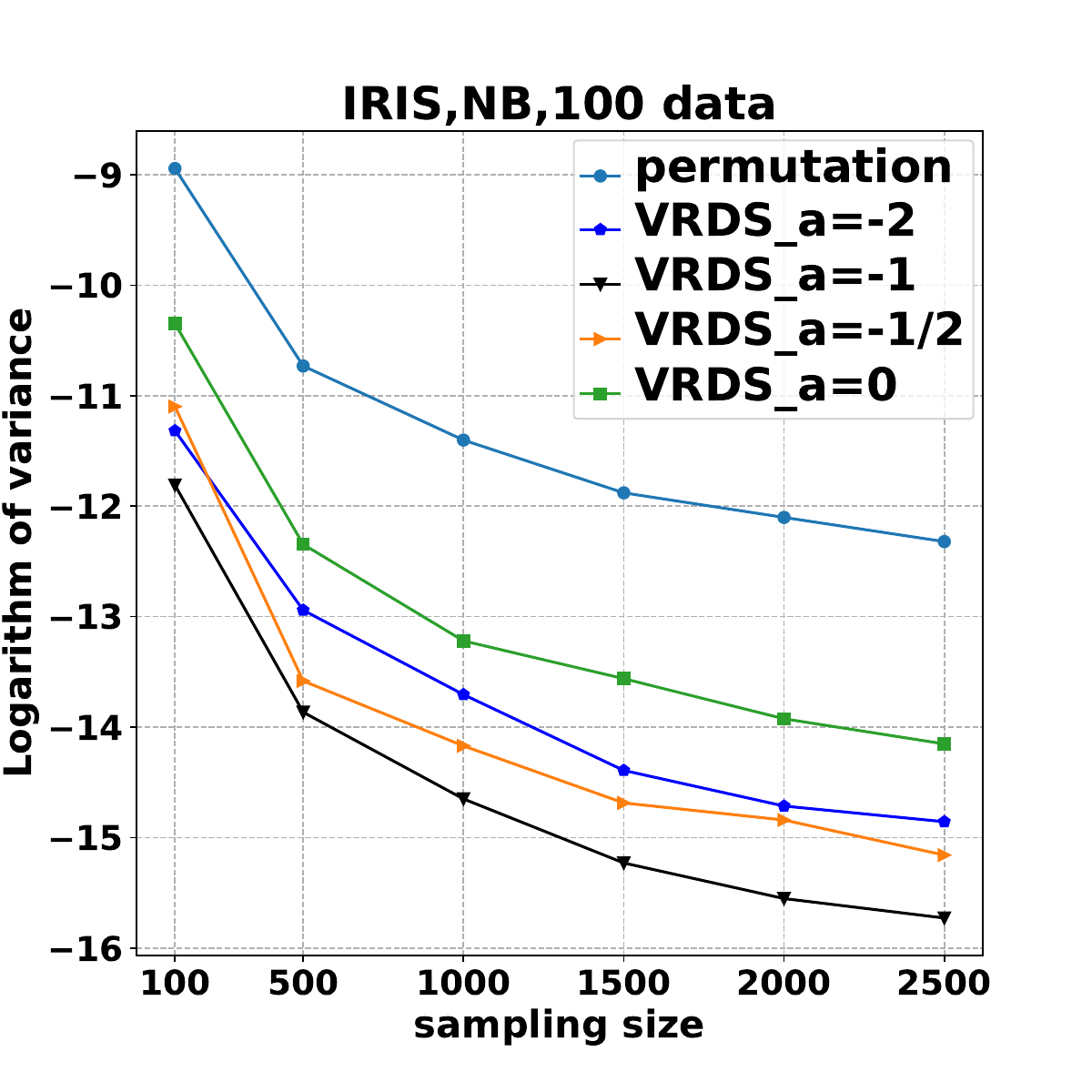}
	\end{minipage}
	\begin{minipage}{0.24\linewidth}
		\centering
		\includegraphics[width=1\linewidth]{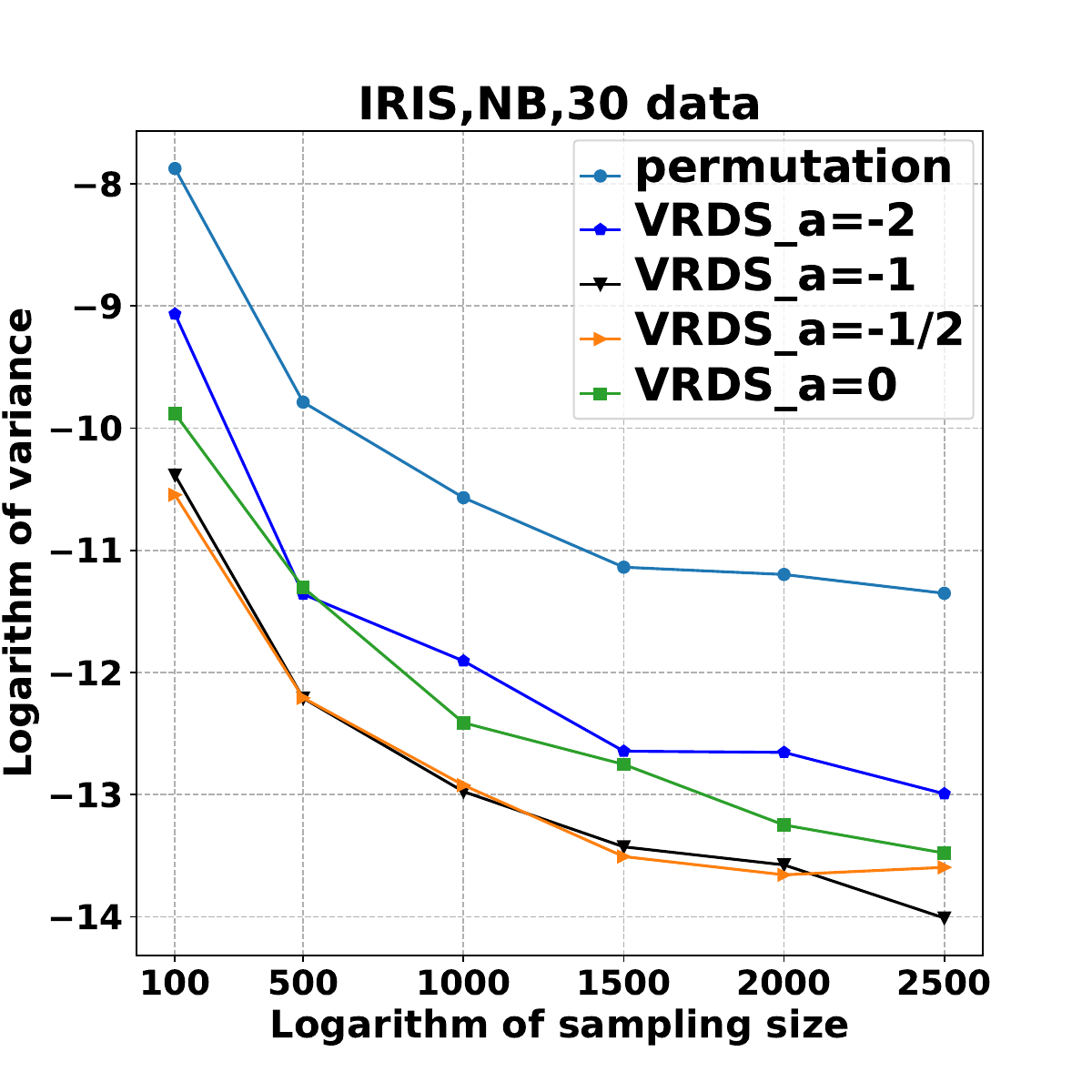}
	\end{minipage}
	
	\begin{minipage}{0.24\linewidth}
		\centering
		\includegraphics[width=1\linewidth]{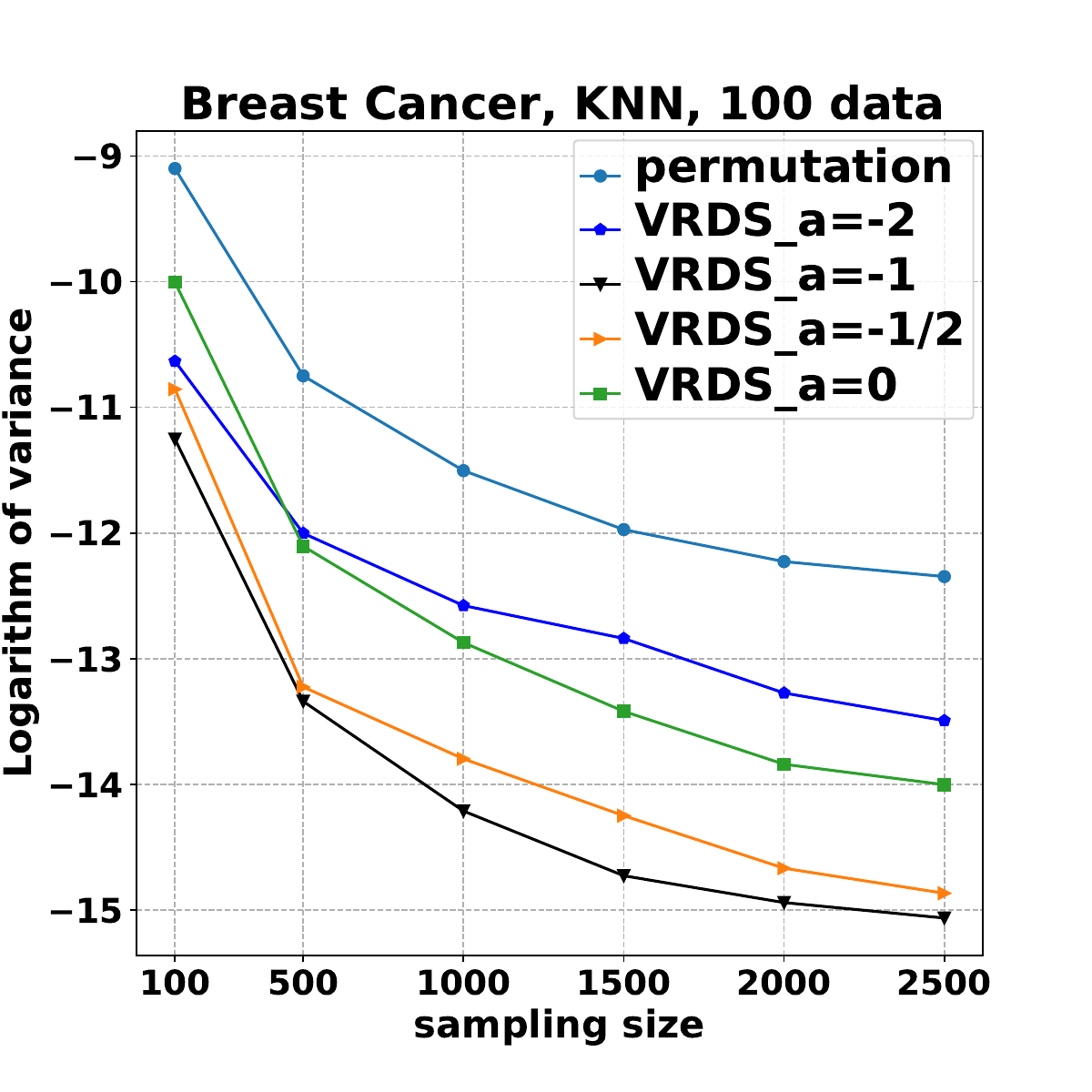}
	\end{minipage}
	\begin{minipage}{0.24\linewidth}
		\centering
		\includegraphics[width=1\linewidth]{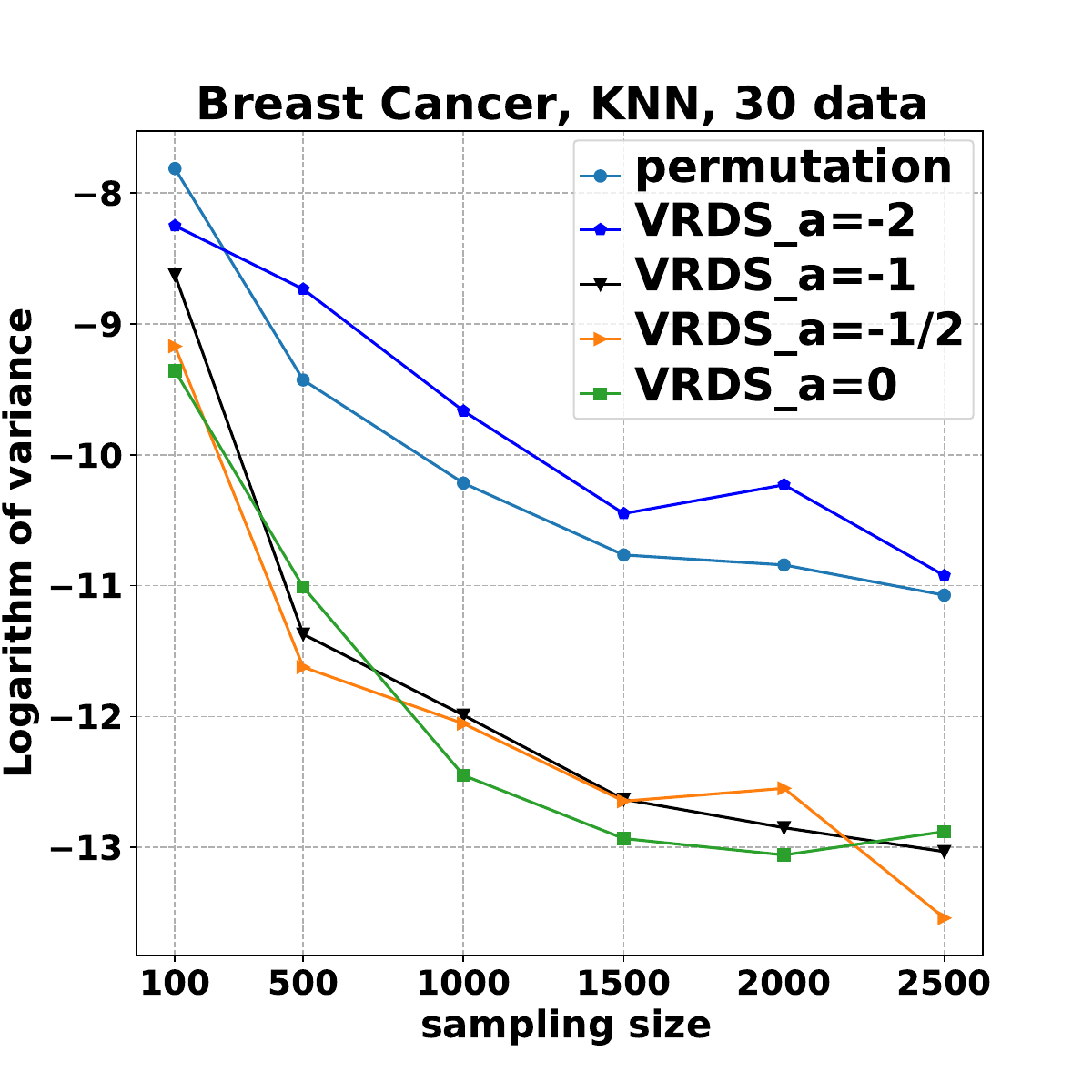}
	\end{minipage}
	\begin{minipage}{0.24\linewidth}
		\centering
		\includegraphics[width=1\linewidth]{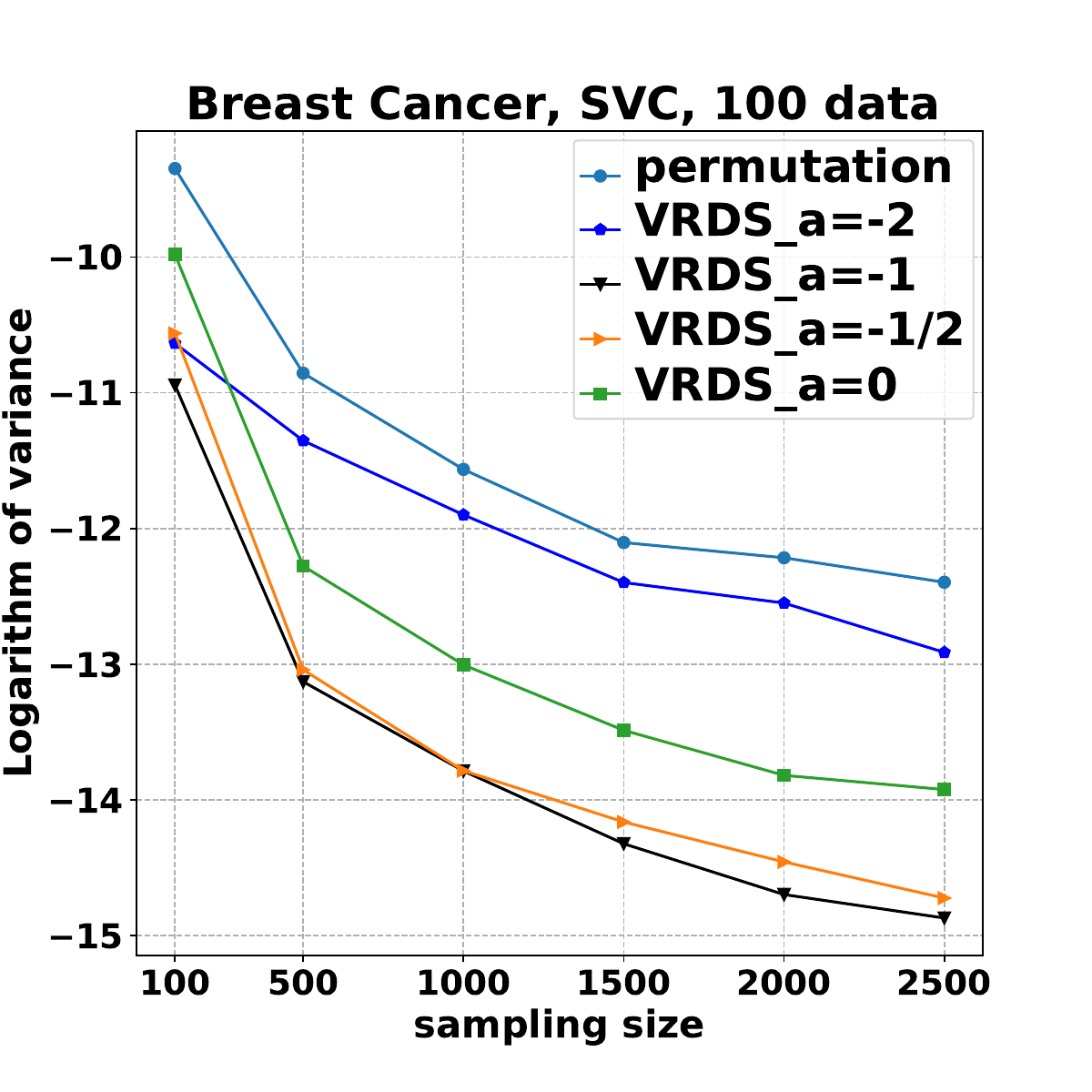}
	\end{minipage}
	\begin{minipage}{0.24\linewidth}
		\centering
		\includegraphics[width=1\linewidth]{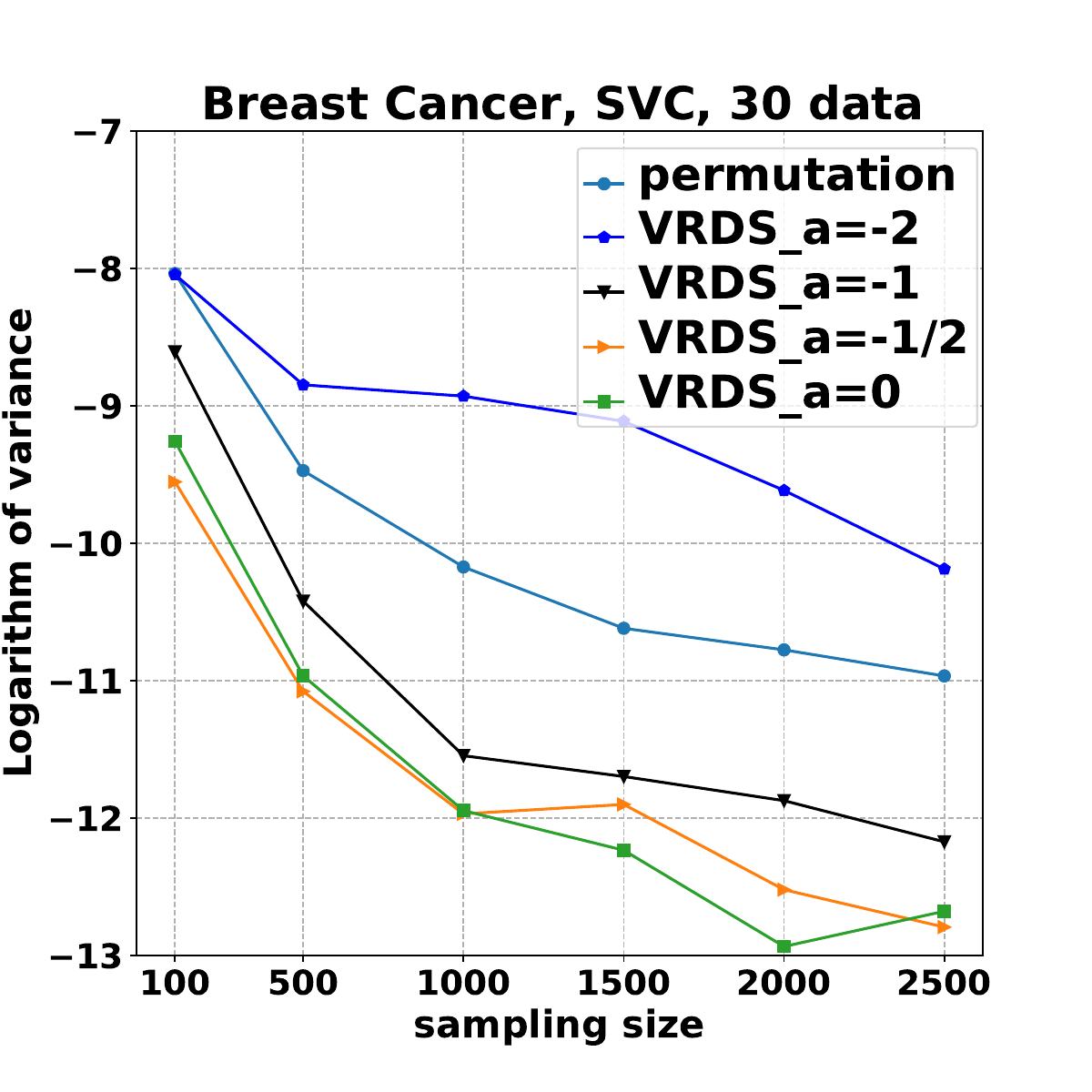}
	\end{minipage}
	
	\begin{minipage}{0.24\linewidth}
		\centering
		\includegraphics[width=1\linewidth]{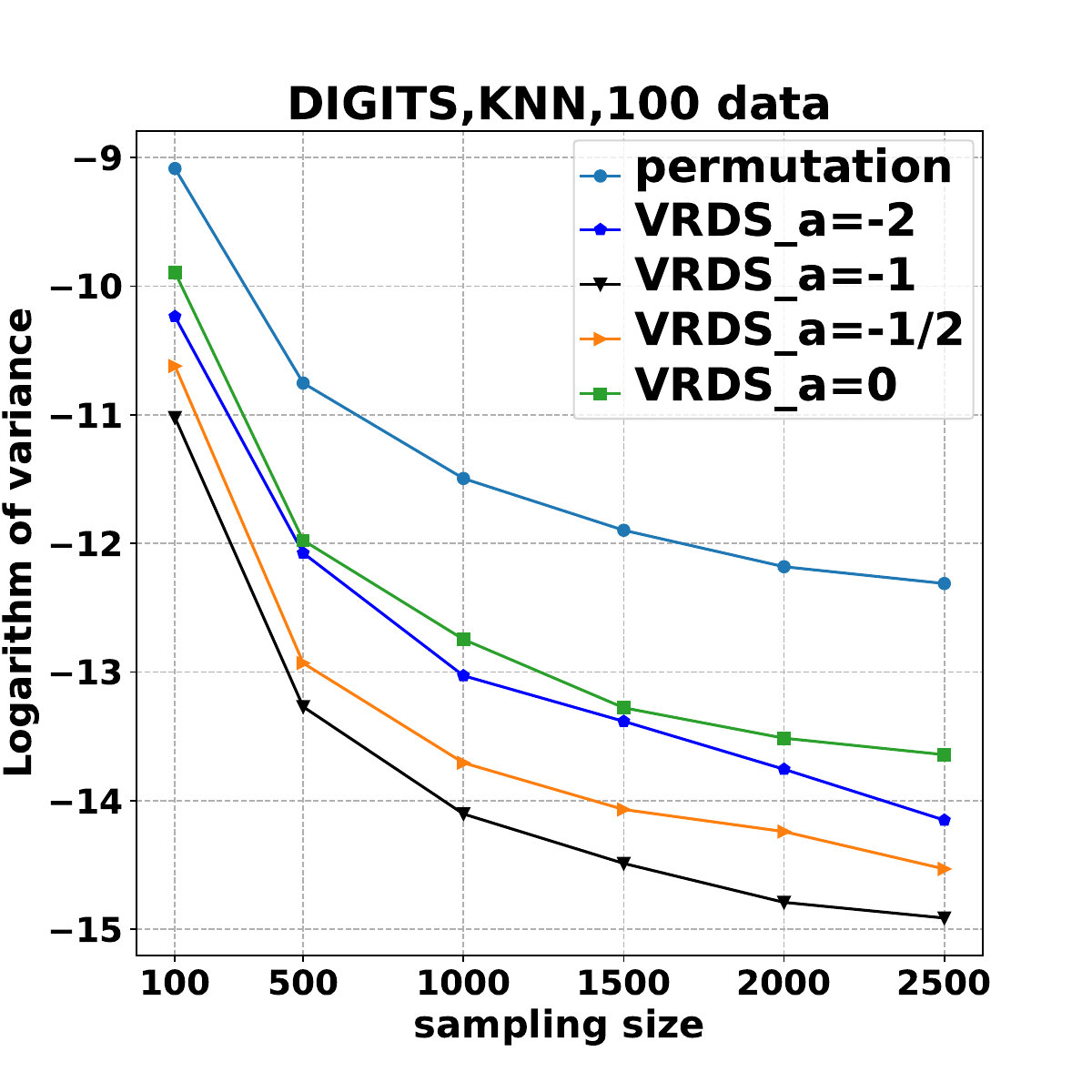}
	\end{minipage}
	\begin{minipage}{0.24\linewidth}
		\centering
		\includegraphics[width=1\linewidth]{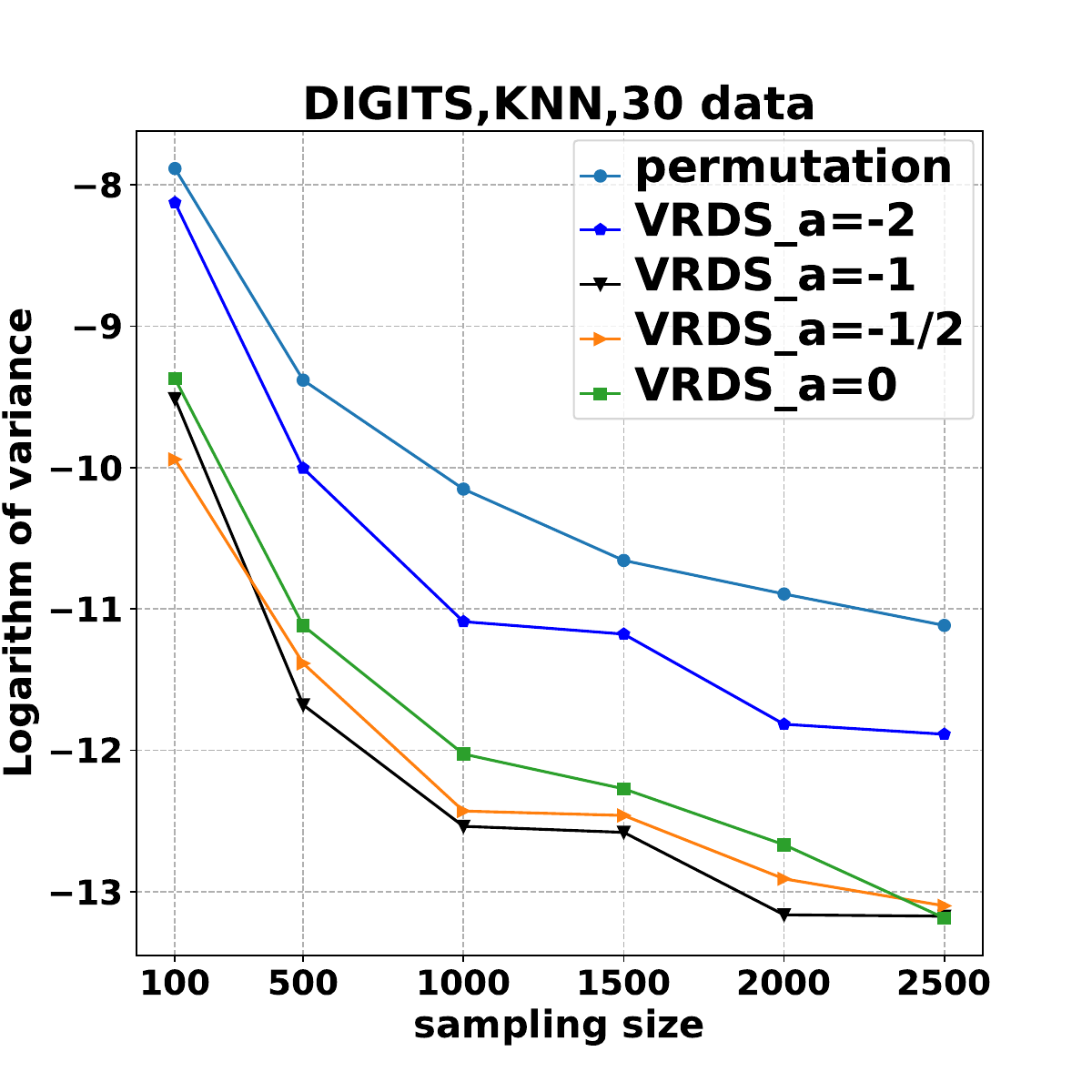}
	\end{minipage}
	\begin{minipage}{0.24\linewidth}
		\centering
		\includegraphics[width=1\linewidth]{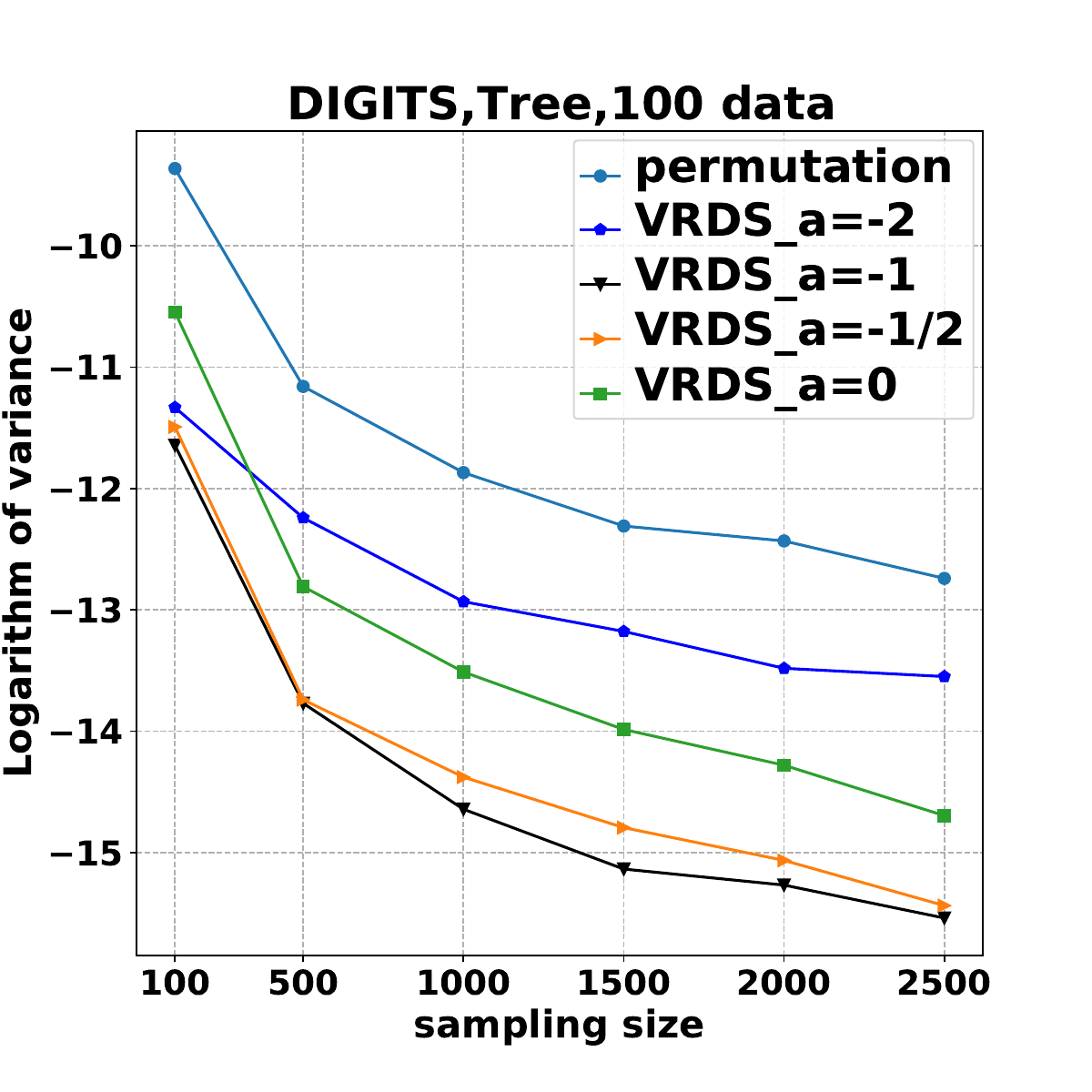}
	\end{minipage}
	\begin{minipage}{0.24\linewidth}
		\centering
		\includegraphics[width=1\linewidth]{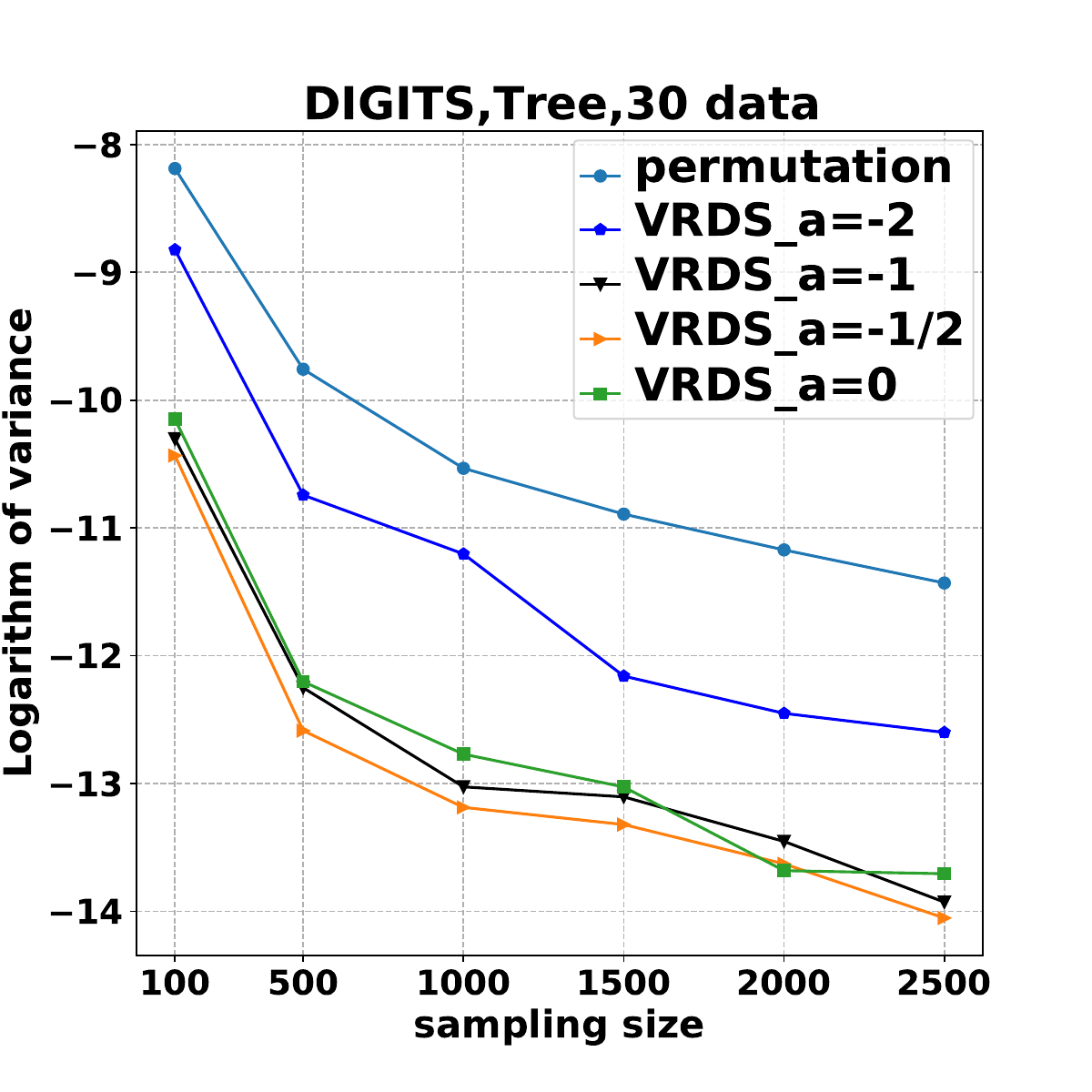}
	\end{minipage}
\caption{Variance of estimated data Shapley on Iris, Breast Cancer and Digits datasets. We use KNN and NB on the Iris dataset, KNN and SVC on the Breast cancer dataset, and KNN and Tree on the digits dataset. Data size is 100 and 30 respectively.}
\label{other datasets}
\end{figure}

 In the part of the parameter selection experiment, we carry out experiments on different datasets, different algorithms and different data sizes.
 Figure \ref{other datasets} shows the same experiment on Iris, Breast cancer and Digits. 
We use KNN and NB on the Iris dataset, KNN and SVC on the Breast cancer dataset, and KNN and Tree on the Digits dataset.
The number of samples varies from 100 to 2500.
It can be seen that $a\in [-1,-1/2]$ performs well in almost all cases.

\begin{figure}[htbp]
	\centering
	\begin{minipage}{0.4\linewidth}
		\centering
		\includegraphics[width=1\linewidth]{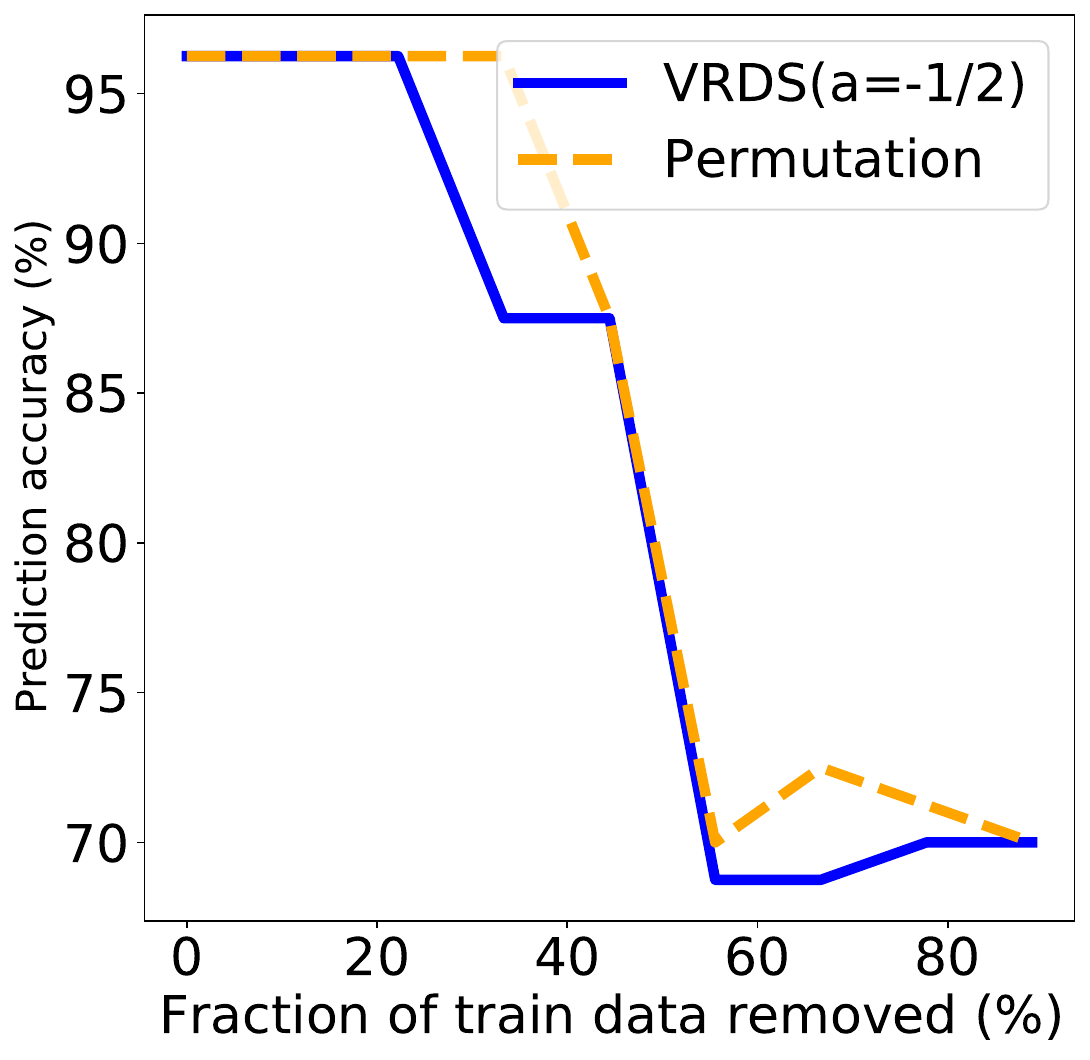}
		\subcaption{}
        \label{drd1}
	\end{minipage}
	\begin{minipage}{0.4\linewidth}
		\centering
		\includegraphics[width=1\linewidth]{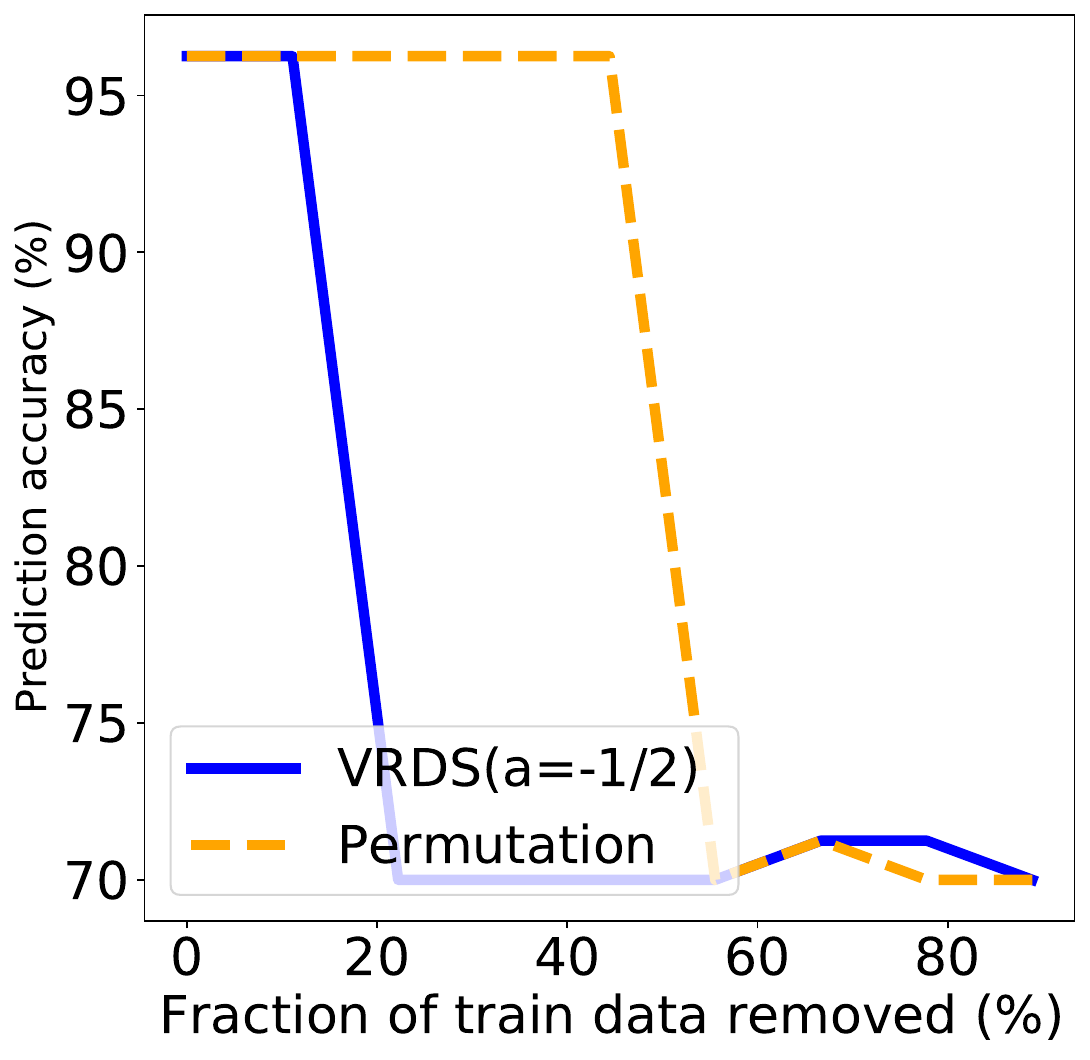}
        \subcaption{}
        \label{drd2}
	\end{minipage}
\caption{Data group removal experiment on the Digits data set. There are $10$ groups and each group has different volume data. VRDS ($a = - \frac{1}{2}$) and permutation sampling algorithms are used to calculate the value of the data group.
 Figure \ref{drd1} shows the removal from the most valuable data group to the least valuable data group. 
 Figure \ref{drd2} shows data removal according to the variance value from large to small.}
\label{data_removal_digits}
\end{figure}

We also carry out experiments on other datasets in the data removal experiment.
 Figure \ref{data_removal_digits} shows the data group removal experiment on the Digits data set. 
 There are $10$ groups, and each group has a different amount of data.
 The minimum group has 5 data and the maximum group has 50 data.
 VRDS with parameter $a = - \frac{1}{2}$ and permutation sampling algorithms are used to calculate the value of the data groups.
 Figure \ref{drd1} shows the removal from the most valuable data group to the least valuable data group. It can be seen the performance of VRDS ($a = - \frac{1}{2}$) decreases faster than that of permutation sampling, which indicates that VRDS ($a = - \frac{1}{2}$) recognize most valuable data earlier.
 Figure \ref{drd2} shows data removal according to the variance value from large to small. 
 Similarly, the prediction accuracy of the VRDS ($a = - \frac{1}{2}$) algorithm decreases faster.

\end{document}